\documentclass[journal]{IEEEtran}
\IEEEoverridecommandlockouts
\usepackage{subcaption}
\usepackage{amsmath} % assumes amsmath package installed
\usepackage{amssymb}  % assumes amsmath package installed
\usepackage{amsfonts}
\usepackage{bm}

\usepackage{amsthm}
\usepackage{mathtools}
\usepackage{hyperref}
\usepackage{algorithmic}
\usepackage{graphicx}
\usepackage{textcomp}
\usepackage{booktabs}
\usepackage{xcolor}
\usepackage{cite}
\usepackage{array}
\usepackage[compact]{titlesec}

\allowdisplaybreaks
\title{Delay-aware Robust Control for Safe Autonomous Driving and Racing}
\begin{document}
%\addtolength{\topskip}{-5mm}
% \addtolength{\topmargin}{-3mm}

% \title{Delay-aware Robust Control for Safe Autonomous Driving\\
% %{\footnotesize \textsuperscript{*}Note: Sub-titles are not captured in Xplore and should not be used}
% %\thanks{Identify applicable funding agency here. If none, delete this.}
% }

\author{Dvij Kalaria$^{1}$, Qin Lin$^{2*}$, and John M. Dolan$^{3}$% <-this % stops a space
%\thanks{*This work was not supported by any organization}% <-this % stops a space
\thanks{$^{1}$Dvij Kalaria is with the Department of Computer Science and Engineering, IIT Kharagpur, India {\tt\small dvij@iitkgp.ac.in}}%
\thanks{$^{2}$Qin Lin is with the EECS Department, Cleveland State University {\tt\small q.lin80@csuohio.edu}}
\thanks{$^{3}$John M. Dolan is with the Robotics Institute, Carnegie Mellon University {\tt\small jdolan@andrew.cmu.edu}}%
\thanks{$^{*}$Corresponding author}
}

\markboth{IEEE Transactions on Intelligent Vehicles}%
{Shell \MakeLowercase{\textit{et al.}}: Bare Demo of IEEEtran.cls for IEEE Journals}

\maketitle

\begin{abstract}
Delays endanger safety of autonomous systems operating in a rapidly changing environment, such as nondeterministic surrounding traffic participants in autonomous driving and high-speed racing. Unfortunately, delays are typically not considered during the conventional controller design or learning-enabled controller training phases prior to deployment in the physical world. In this paper, the computation delay from nonlinear optimization for motion planning and control, as well as other unavoidable delays caused by actuators, are addressed systematically and unifiedly. To deal with all these delays, in our framework: 1) we propose a new filtering approach with no prior knowledge of dynamics and disturbance distribution to adaptively and safely estimate the time-variant computation delay; 2) we model actuation dynamics for steering delay;  3) all the constrained optimization is realized in a robust tube model predictive controller. For the application merits, we demonstrate that our approach is suitable for both autonomous driving and autonomous racing. Our approach is a novel design for a standalone delay compensation controller. In addition, in the case that a learning-enabled controller assuming no delay works as a primary controller, our approach serves as the primary controller's safety guard. The video demonstration is available online \footnote[1]{\href{https://youtu.be/bWNpzMi3RlI}{https://youtu.be/bWNpzMi3RlI}}.

\end{abstract}

\section{Introduction}

The need for high-resolution sensors and powerful onboard computing represent a significant barrier to achieving inexpensive autonomous vehicles. While much has been done to improve the efficiency of algorithms, the required computation time is unavoidable. The motion planning and control component of a self-driving car requires heavy computation in highly complex traffic environments (usually nonlinear and non-convex). Because any non-negligible delay is risky in fast-changing conditions, e.g., highway and autonomous racing, the frequently held premise that the optimal control action is produced instantaneously from an optimization program and executed on the plant is erroneous.
\begin{figure}[htbp] 
    \centering
    \includegraphics[width=0.4\textwidth]{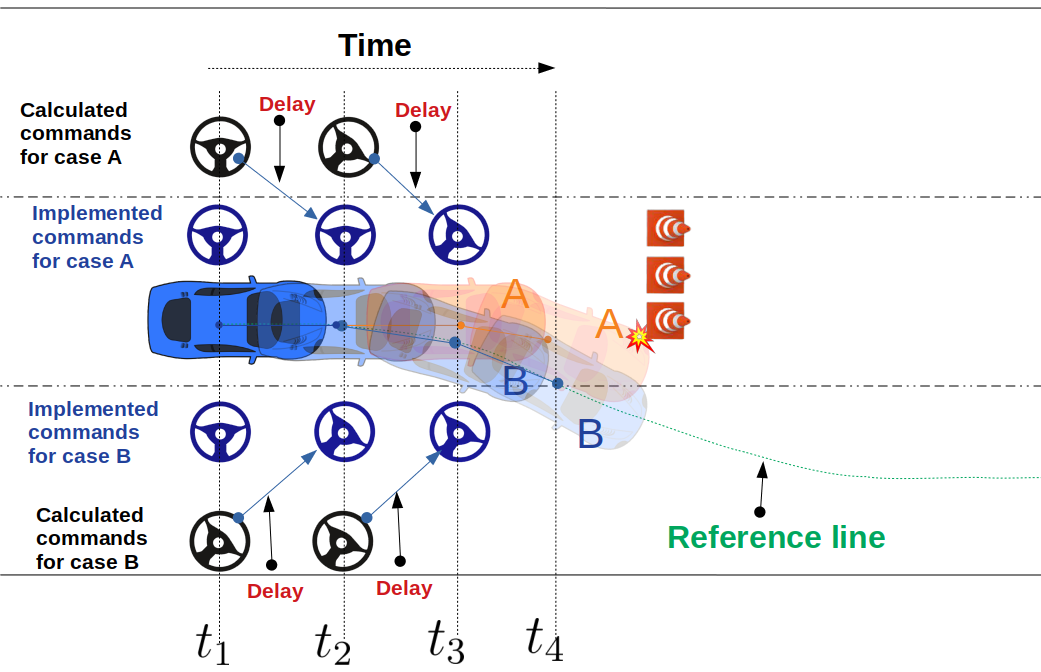}
    \caption{An illustration comparing a conventional control and a delay-aware control. The conventional method, as shown in case A, always uses the current state of the vehicle to compute the future control action. Due to a delay, the calculated action will not be instantaneously executed (see the shifted steering wheel signs from black to blue). Our controller is aware of the delay factor, thus in the earliest time step (see the leftmost calculated command in case B), a steering command has been calculated. The movements in case A with the orange color are from the late steering command sequence shown in the first row of blue steering wheel signs. Clearly, the movements in case B with the blue color are safer.}
    \label{fig:start_fig}
\end{figure}

Fig. \ref{fig:start_fig} illustrates the motivation of our work in a collision avoidance scenario. The sequential movements in orange color result from a traditional controller without considering any delay, which could be hazardous in a highly dynamic environment (e.g., high-speed scenario). However, our delay-aware controller takes a safe action earlier (e.g., earlier steering for an evasive maneuver with the blue color). Our approach  deals with three types of delay in a unified manner: 1) actuator dynamics delay; 2) control action processing delay; and 3) computation time delay. First, we augment the commonly used vehicle dynamic modeling by including steering action delay. Second, we propose a novel adaptive Kalman filter variant: INFLUENCE (adaptIve kalmaN FiLter with Unknown process modEl and Noise CovariancE) to probabilistically and safely estimate the control action processing delay and computation time delay caused by online optimization. INFLUENCE simultaneously identifies an unknown process model and noise covariances. Third, we extend robust tube model predictive control (MPC) theory by adding a new delay-aware feature for optimizing control.

The first controller design is proposed as a standalone delay-aware robust control framework, while the other design aims at compensating delays for an existing learning-enabled (LE) controller for a provable safety guarantee, which aligns with the safe artificial intelligence (AI) theme. Naive delay-free assumptions are frequently made when the LE control policy is trained in a simulation environment. Such a simulation-to-real (sim-to-real) gap hinders reliable deployment of safety-critical autonomous systems like self-driving cars. The LE controller is treated as a high-level reference and allows a low-level refinement. We use delay compensation to actively regulate its unsafe control. We validate our framework in two challenging scenarios: the first one is  on-road autonomous driving in a dynamic and uncertain traffic environment. The surrounding traffic participants have non-deterministic behaviors such as unexpected braking. The second is autonomous racing, which can reach high speed up to 180 km/h.

We make the following contributions:

\begin{enumerate}
    \item We extend the conventional robust tube MPC framework with the new capacity to deal with various types of practical delays.
    \item Unlike many existing filters, INFLUENCE does not require prior knowledge of model dynamics and noise distribution. In addition, its probabilistic and adaptive estimation mitigates overconservatism.
    \item We propose a safety guard component to compensate for delays of a LE controller without considering any delays. In addition, in our control optimization framework, the delay compensator can be combined with barrier functions (CBFs) for collision avoidance.
\end{enumerate}

% 1. A unified delay-aware robust control approach dealing with three major delays: computation time delay, actuator command processing delay, and actuator dynamics delay

% 2. INFLUENCE is a probabilistic framework for real-time estimation of computation time instead of taking an upper bound, which makes the algorithm safe while minimizing conservativeness. INFLUENCE has application merit for general safe prediction problems, since it does not assume known process model and noise covariance.

% 3. A control plan to safely compensate for these delays for a LE controller which doesn't consider delays while also safeguarding the LE controller against collision avoidance using control barrier functions (CBFs).

% 4. Use of racing line as the frenet frame reference to leverage independence from complex road geometry especially necessary for autonomous racing.

We have the following two substantial improvements over the preliminary version \cite{dvij2022}: 1) In this study, we use a dynamic vehicle model rather than a kinematic model. We demonstrate the scalability of our algorithm with a higher state dimension. 2) Collision avoidance is not taken into account in the compensation control of a learning-enabled controller in \cite{dvij2022}. In our current work, we combine our control strategy with an MPC embedded with CBFs. The novel integration can be called a delay-aware robust quadratic programming with control barrier functions (DR-QP-CBFs).

The rest of this paper is organized into the following sections. A review of some relevant related works is presented in Section \ref{sec:related_work}. The methodology is described in Section \ref{sec:method}. The experiments and their results are presented in Section \ref{sec:experiment}. Section \ref{sec:conclusions} contains the conclusions.

\section{Related work}
\label{sec:related_work}
%Autonomous driving is an active field of research that poses challenges in various domains, including Motion Planning, Controls, and Perception.
%Motion planning in the context of autonomous driving has been a widely studied and active research topic. The current approaches to motion planning for ground vehicles can be categorized into 1) graph search-based methods; 2) sampling-based methods; 3) curve interpolation-based methods; and 4) optimization-based approaches using sequential programming. All these approaches are well reviewed in \cite{gonzalez2015review}.

%However, in most methods, even though safety is guaranteed in the high-level path planner, safety is not assured in the low-level control due to the ``planning and control inconsistency" problem caused by tracking error \cite{polack2017kinematic}. Optimization-based approaches do consider vehicle dynamics in the trajectory optimization; however, due to its high computation cost, it is still impossible to consider sophisticated vehicle dynamic models. To address this problem, our recent work \cite{khaitan2020safe} considers collision avoidance by efficient addition of convex state constraints in the MPC algorithm used for tracking.

%A popular architecture for layered planning and control involves a high-level planner and a low-level controller for trajectory tracking such as Ackermann controller \cite{din2019real}, pure pursuit \cite{conlter1992implementation}, Stanley \cite{rokonuzzaman2021review}, Model Predictive Control (MPC) \cite{borrelli2005mpc}, and preview controller \cite{liao2018design} to name a few.

In this section, we will briefly review the most important related work on how to deal with delays and some CBFs work for safe control.

\subsubsection{General delay compensation control}
\cite{cortes2011delay} considers a discrete control problem: it proposes to shift one discrete step ahead for the initial state in MPC, for the particularly simple one-step delay problem. \cite{su2013computation} uses a cache mechanism for buffering previously computed control actions. However, their approach only works for static scenarios with fixed horizon length, which leads to a less responsive controller. The first significant improvement in our work is that our filtering approach actively estimates a time-variant local upper bound of the computation delay.

\subsubsection{Actuation delay compensation control}
To deal with delays in actuator dynamics, \cite{nahidi2019study} proposes to augment the original state space model with an extra first-order ordinary differential equation (ODE) for the delay behavior. The particular application in their work is braking control. For control action processing delay, which is usually time-invariant, \cite{liao2018design} proposes to compensate for such a delay by transitioning the initial state in a preview controller. \cite{6083022} further extends the idea and considers actuator saturation. The major disadvantage of the preview controller is that it is not flexible enough to simultaneously consider multiple constraints such as dynamics, state, and control limits. Note that our work deals with delays and all constraints of state and control in a unified robust MPC optimization.

\subsubsection{Provable safety using CBFs}
 Wieland and Allower \cite{Peter2007} in 2007 first proposed the CBF-based method which describes an admissible control space capable of rendering forward invariance of a safe set. Later, Aaron et al. \cite{Amescdc2014} extended CBF to a minimally restrictive setting and applied it to the lane-keeping and adaptive cruise control problem.

\cite{orosz2019safety} presents a unique proof for the stability and safety of time delay state systems. However, it does not account for control systems. Later, \cite{singletary2020control} proposed safety of control barrier functions for sampled-data systems where the control signal is updated in discrete steps as regular time intervals but the system is evolved continuously. Such systems are close to real systems, as we can only update the control signal in discrete steps. It proves validity of control barrier functions under such a setting. It later exploits the formulation to extend the proof to systems with constant known input delays. However, in our case we require validity for time-varying input delay. \cite{abel2021safety} describes an approach for designing safety-critical controllers involving control barrier functions for nonlinear systems with time-varying input delay. It combines a state predictor that compensates for the time-varying input delay with a CBF-based feedback law for the nominal system, which is delay-free. However, this is for continuous systems with continuous control. In our case, the system is discretely controllable with variable delay time. Also, we use the CBF accounting for variable input delay in conjunction with MPC to account for uncertainties in the system model and environment due to dynamic obstacles.  
\section{Methodology}
\label{sec:method}

\subsection{Notation}
A polytope is the convex hull of finite points in Euclidean $d$-dimensional space with vertices of the form $\mathbb{R}^d$.
The Minkowski sum of two polytopes, $A$ and $B$, is another polytope in $d$-dimensional space defined as $A \oplus B := \{a + b | a \in A, b \in B\}$.
The Pontryagin difference between two polytopes, $A$ and $B$, is another polytope in $d$-dimensional space defined as $A \ominus B := \{x | x + b \in A, \forall b \in B\}$.

\subsection{Optimal racing line} \label{opt_racing_line}

During a race, professional drivers follow a racing line while using specific maneuvers that allow them to use the limits of the car’s tire forces. This path can be used as a reference by the autonomous racing motion planner to effectively follow time-optimal trajectories while avoiding collision, among other objectives. Many recent works have proposed motion planners on top of optimal racing lines including sampling-based local planners \cite{feraco2020local} \cite{stahl2019} \cite{tramacere2021local} and optimization-based approaches \cite{Srinivasan2021AHM} \cite{Vzquez2020OptimizationBasedHM} \cite{Kalaria2021LocalNO}. The racing line is essentially a minimum-time path or a minimum-curvature path. They are similar but the minimum-curvature path additionally allows the highest cornering speeds given the maximum legitimate lateral acceleration \cite{doi:10.1080/00423114.2019.1631455}. There are many proposed solutions to finding the optimal racing line, including dynamic programming \cite{Beltman2008OptimizationOI}, nonlinear optimization \cite{Rosolia2020LearningHT} \cite{doi:10.1080/00423114.2019.1631455}, genetic algorithm-based search \cite{Vesel2015RacingLO} and Bayesian optimization \cite{Jain2020ComputingTR}. However, for our work, we calculate the minimum-curvature optimal line, which is close to the optimal racing line as proposed by \cite{doi:10.1080/00423114.2019.1631455}. The race track information is input by a sequence of tuples ($x_i$,$y_i$,$w_i$), $i \in \{0,...,N-1\}$, where ($x_i$,$y_i$) denotes the coordinate of the center location and $w_i$ denotes the lane width at the $i^{th}$ point, vehicle width $w_{veh}$. The output trajectory consists of a tuple of seven variables: coordinates $x$ and $y$, curvilinear longitudinal displacement $s$, longitudinal velocity $v_x$, acceleration $a_x$, heading angle $\psi$, and curvature $\kappa$. The trajectory is obtained by minimizing the following cost:
\vspace{-0.2cm}
\begin{equation} \label{opt_racing_line_eqn}
\begin{split}%\label{total cost term}
    & \mathop{\min}\limits_{\alpha_1...\alpha_N} \quad \sum_{n=0}^{N-1} \kappa_i^2(n)\\
    \text{s.t.} & \quad \alpha_i \in \left[ -w_i+\frac{w_{veh}}{2},w_i-\frac{w_{veh}}{2} \right]\\
\end{split}
\end{equation}
where $\alpha_i$ is the lateral displacement at the $i^{th}$ position. To generate a velocity profile, the vehicle's velocity dependant longitudinal and lateral acceleration limits are required  \cite{doi:10.1080/00423114.2019.1631455}. Using the optimal racing line as a reference, we use a sampling-based motion planner in the Frenet frame \cite{Werling2010OptimalTG}. %More experimental details can be found in \cite{khaitan2021safe}. %Readers are referred to  for more details.

\subsection{Frenet motion planner} \label{frenet_motion_planner}
 The longitudinal and lateral motion are treated differently and the position is represented by the longitudinal displacement $s(t)$ and the lateral displacement $d(t)$ obtained from the perpendicular to the global position ($x(t)$,$y(t)$) on the racing line. %The process of generating trajectories can be described as a curve interpolation method which combines a quartic polynomial for longitudinal motion and a quintic polynomial for lateral motion to generate an optimal collision-free trajectory. It weighs the generated
We define a cost-function ($C$) as:

\begin{equation} \label{frenet_planner_cost}
\begin{split}%\label{total cost term}
C &= k_{lat} C_{lat} + k_{lon} C_{lon} + k_{obs} C_{obs}\\
C_{lat} &= k_j\dddot{d} + k_a \ddot{d} + k_v \dot{d}\\
C_{lon} &= k_j\dddot{s} + k_a \ddot{s} + k_v \dot{s}\\
C_{obs} &= \sum_{k=0}^{N-1} \frac{1}{\sqrt{(s_k-s_{obs,k})^2+(d_k-d_{obs,k})^2}}
\end{split}
\end{equation}
% \vspace{-0.2cm}
where $C_{obs}$ is the sum of inverse distances from the nearest obstacle at each time step. $k_{lat}$, $k_{lon}$, and $k_{obs}$ are respectively cost weights for lateral motion, longitudinal motion, and obstacle avoidance. $k_j$, $k_a$, and $k_v$ are respectively cost weights for jerk, acceleration, and velocity. The optimal planned trajectory is the one with the minimum cost among all sampled trajectories. For dynamic obstacle avoidance, each trajectory point is checked for an overlap with the predicted obstacle positions at the corresponding time. The future obstacle positions ($s_{obst,k}$,$d_{obst,k}$) are predicted by assuming constant longitudinal and lateral velocity of the obstacle. However, this can be replaced by more sophisticated predictors \cite{Deo2018ConvolutionalSP,pan2020safe,khaitan2021safe}.

\subsection{System dynamics} \label{actuator_dynamics}
A dynamic bicycle model is used to describe the dynamics of the vehicle, see Eq. \ref{eq:dynamic_eqn}. For more details and derivation, readers are referred to Chapters 2 and 3 from \cite{book}.
\begin{equation}
\label{eq:dynamic_eqn}
    \begin{split}
        \begin{bmatrix}
                \dot e_1 \\
                \ddot e_1 \\
                \dot e_2 \\ 
                \ddot e_2\\
                \dot v_x\\
                \dot s\\
                \dot \delta_{a}
        \end{bmatrix}	
                &= 
        \begin{bmatrix}
            \dot e_1 \\
            -\frac{2 C_f + 2 C_r}{m V_x} \dot e_1 + \frac{2 C_f + 2 C_r}{m} e_2 - \frac{-2 C_f l_f + 2 C_r l_r}{m V_x} \dot e_2 \\
            \dot e_2 \\
            -\frac{2 C_f l_f - 2 C_r l_r}{I_z V_x} \dot e_1 + \frac{2 C_f l_f - 2 C_r l_r}{I_z} e_2 - \frac{2 C_f l_f^2 + 2 C_r l_r^2}{I_z V_x} \dot e_2 \\
            - K_f v_x\\
            v_x - \dot e_1 e_2\\
            -K_{\delta} \delta_a
        \end{bmatrix}\\
        &+
        \begin{bmatrix}
            0 \\
            - v_x (v_x + \frac{2 C_f l_f - 2 C_r l_r}{m v_x})\\
            0 \\
            - \frac{2 C_f l_f^2 + 2 C_r l_r^2}{I_z}\\
            0\\
            0\\
            0
        \end{bmatrix} c
        +
        \begin{bmatrix}
            0 \\
            2 \frac{2 C_f}{m} \delta_a\\
            0 \\
            0 \\
            K_p p\\
            0\\
            K_{\delta} \delta 
        \end{bmatrix}\\
        \mathbf{u} &= \begin{bmatrix} \delta & p \end{bmatrix}^T\\
        \mathbf{x} &= \begin{bmatrix} e_1 & \dot e_1 & e_2 & \dot e_2 & v_x & s & \delta_a \end{bmatrix}^T
    \end{split}
\end{equation}
% \qin{what does $c$ mean in the above equation?} 
where $C_f$, $C_r$ are the stiffness coefficients of the front and the rear tires, $l_f$ and $l_r$ are the front and the rear tire distances from the centre of mass (COM), $I_z$, $m$ are the moment of inertia and mass of the vehicles, and $c$ is the inverse of the radius of curvature of the reference line at the perpendicular point from the vehicle's position. The control variables are steering angle ($\delta$) and pedal ($p$). The state variables taken include the lateral position error ($e_1$), heading angle error ($e_2$) with respect to the optimal racing line, their first-order derivatives $\dot{e}_1$ and $\dot{e}_2$, longitudinal velocity ($v_x$), and longitudinal displacement ($s$). In many existing motion planning and control works, steering angle $\delta$ and acceleration are assumed to be applied instantaneously in an ideal case. Due to this, a delay is caused due to the mismatch between the calculated and the actual steering angle state. To solve this problem, we include the steering angle as a separate state, and model the steering actuator dynamics as follows by a first order ordinary differential equation (see \cite{nahidi2019study}): $\Dot{\delta_a} = K_{\delta} (\delta - \delta_a)$, where $\delta$ is the desired steering angle, $\delta_a$ is the actual steering angle, and $K_{\delta}$ is calculated as the inverse of the time constant. Thus, we control only $\delta$ to tell the actuator which state to achieve and use $\delta_a$ instead of $\delta$ for the dynamics calculation. For acceleration, the delays in pedal dynamics are neglected for simplicity. $K_p$ is used for the mapping between pedal and acceleration. However, its delay can also be modelled in the same way as the steering.
% We include the actual steering angle as a state denoted as $\delta_{a}$ in the system dynamics. The desired steering angle $\delta$ is the actual control command. Inspired by the spirit in \cite{nahidi2019study}, we model the change in the steering angle state by a first-order ODE, i.e., $\Dot{\delta_a} = K_{\delta} (\delta - \delta_a)$, where $K_{\delta}$ is the inverse of the time constant. For acceleration, the delays in pedal dynamics are neglected for simplicity. $K_p$ is used for the mapping between pedal and acceleration. However, its delay can also be modelled in the same way as the steering. %The continuous dynamic model is discretized in our experiments. The vehicle dynamics are approximately discretized by assuming constant state derivative for a short period of time $dt$ as given in \ref{eq:dynamic_eqn}.

\subsection{Control barrier functions}

CBFs ensure safety by rendering a forward-invariant safe set. We define a continuous and differentiable safety function $h(\bm{x}): \mathcal{X} \xrightarrow{} \mathbb{R}$. The \emph{superlevel set} $\mathcal{C} \in \mathbb{R}^n$ can be named as a safe set. Let the set $\mathcal{C}$ obey
\begin{align}
    \mathcal{C} = \{\mathbf{x} \in \mathcal{X}: h(\mathbf{x})\geq 0 \} \\
    \label{eq:C}
    \partial \mathcal{C} = \{\mathbf{x} \in \mathcal{X}: h(\mathbf{x})=0 \} \\
    \text{Int}( \mathcal{C}) = \{\mathbf{x} \in \mathcal{X}: h(\mathbf{x})>0 \}.
\end{align}

A control affine system has the form $\dot{\mathbf{x}} = f(\mathbf{x})+g(x)\mathbf{u}$ like Eq. \ref{eq:dynamic_eqn}, such that
\begin{equation}
    \exists u \quad \text{s.t.} \quad \dot{h}(\mathbf{x}) \geq -\kappa_h (h(\mathbf{x}))
\end{equation}
where $\kappa_h \in \mathcal{K}$ is particularly chosen as $\kappa_h(a) = \gamma a$ for a constant $\gamma > 0$. Also, the time derivative of $h$ is expressed as

\begin{equation}
    \dot{h}(\mathbf{x}) = L_fh(\mathbf{x}) + L_g h(\mathbf{x}) \mathbf{u}.
    \label{eq:hdot}
\end{equation}
where $L_f h(\mathbf{x})$ and $L_g h(\mathbf{x})$ represent the Lie derivatives of the system denoted as $\nabla h(\mathbf{x}) f(\mathbf{x})$ and $\nabla h(\mathbf{x}) g(\mathbf{x})$, respectively. The safety constraint for a CBF is that there exists a $\gamma >0$ such that

\begin{equation}
    \underset{\mathbf{u} \in \mathcal{U}}{\inf}(L_f h(\mathbf{x}) + L_g h(\mathbf{x}) \mathbf{u} ) \geq -\gamma (h(\mathbf{x}))
    \label{cbf}
\end{equation}
for all $\mathbf{x} \in \mathcal{X}$. The solution $\mathbf{u}$ assures that the set $\mathcal{C}$ is a forward invariant, i.e.,  $x(t \xrightarrow{} \infty) \in \mathcal{C}$. In a practical collision avoidance task, the safety function can be designed as the relative distance between the ego system and a dynamic obstacle.

Recently, CBFs are used in conjunction with MPC  \cite{Zeng2021SafetyCriticalMP,WILLS20041415}. Having a barrier function-based constraint in MPC not only ensures strict satisfaction of safety constraints, but also facilitates a smooth trajectory. 

% \subsubsection{HOBCBF}

\subsection{Robust Tube MPC}

The uncertain plant is assumed to be described by the discrete-time system model as $x_{k+1}=f(x_k,u_k,w_k)$. The disturbance $w_k$ can be used to describe model uncertainties and external disturbance from the environment. Considering function $f$ to be linear, we can describe the dynamics as $x_{k+1} = A x_k + B u_k + w_k$ with $A\in \mathbb{R}^{n \times n}$ and $B\in \mathbb{R}^{n \times m}$. For a stable feedback system with $A_K = A + BK$, where $K \in \mathbb{R}^{m \times n}$ is the control gain. The disturbance $w_k$ is assumed to be bounded as $w_k \in W$. $W$ is assumed to be a convex polyhedron with origin as its interior point. The objective is to stabilize this uncertain system while satisfying the state and control constraints as $x_k \in X$ and $u_k \in U$. For this system, we can define a disturbance invariant set, $Z$ such that $A_K \mathcal{Z} \oplus \mathcal{W} \subseteq \mathcal{Z}$. A typical MPC controller is defined as a finite horizon optimization problem, $P$ over the future input control sequence and states. We define $P$ on equivalent nominal system assuming no disturbance to optimize nominal state sequence $\Bar{X} = \{\Bar{\bm{x}}_0,\Bar{\bm{x}}_1...,\Bar{\bm{x}}_N\}$, nominal control sequence $\Bar{U} = \{\Bar{\bm{u}}_0,\Bar{\bm{u}}_1...,\Bar{\bm{u}}_{N-1}\}$ as follows,
\begin{equation} \label{objective function}
\begin{split}%\label{total cost term}
    \mathop{\min}\limits_{U}&\quad\sum_{t=0}^{N-1} (\Bar{\bm{x}}_k-{\bm{x}_{ref,k}})^T Q (\Bar{\bm{x}}_k-{\bm{x}_{ref,k}})+\Bar{\bm{u}}_k^T R \Bar{\bm{u}}_k\\&+(\Bar{\bm{x}}_N - {\bm{x}_{ref,N}})^T Q_N (\Bar{\bm{x}}_N - {\bm{x}_{ref,N}})\\
    s.t. &\quad \Bar{\bm{x}}_{k+1} = A \Bar{\bm{x}}_k + B \Bar{\bm{u}}_k\\
    &\quad \bm{x}_{0} \in \Bar{\bm{x}}_0 \oplus \mathcal{Z}\\
    &\quad \Bar{\bm{u}} \in \mathcal{U} \ominus K\mathcal{Z}\\
    &\quad \Bar{\bm{x}} \in \mathcal{X} \ominus \mathcal{Z}
\end{split}
\end{equation}
where $X_{ref} = \{ \bm{x}_{ref,0}, \bm{x}_{ref,1}..., \bm{x}_{ref,N} \}$ is the reference state sequence, $R$, $Q$ and $Q_N$ are the control, state and terminal state cost matrices, respectively and all of them are positive semi-definite. Now, for this uncertain system, if we design a controller to give the following command : $\bm{u} = \Bar{\bm{u}} + K(\bm{x}-\Bar{\bm{x}})$, where $\bm{x}$ is the observed state. For such a controller, we can guarantee that $\bm{x}^+ \in \Bar{\bm{x}}^+ + \mathcal{Z}$ for any bounded $\bm{w} \in \mathcal{W}$ which implies that all states $\bm{x}_k$ will be robustly  contained inside $\mathcal{X}$. In our case, we have a nonlinear model function $f$ in Eq. (\ref{eq:dynamic_eqn}) which we can convert to an equivalent linear time variant (LTV) system with model function $f_k$, where the matrices $A_k$ and $B_k$ are obtained by the Jacobians at the current state, $x_k$. Readers are referred to \cite{1383612} for more details and a detailed proof on the stability and feasibility of this controller. Also, it must be noted that we can inculcate the error in the linearized model w.r.t the actual model $f$ by including an additional disturbance for it as described in \cite{gao2014}. For our work, we assume that the extra disturbance has been enclosed by a sufficiently large disturbance margin.
% where $Q$, $R$ and $Q_N$ are the state, control and terminal state cost matrices, respectively. The controller will give the following command : $\bm{u} = \Bar{\bm{u}} + K(\bm{x}-\Bar{\bm{x}})$, at the current observed state $\bm{x}$. This guarantees $\bm{x}^+ \in \Bar{\bm{x}}^+ + \mathcal{Z}$ for any $\bm{w} \in \mathcal{W}$, i.e., all states $\bm{x}_k$ will be inside the constraint set $\mathcal{X}$. 
% However, in our case, for the nonlinear system, we use the equivalent LTV system. We use Jacobian matrices $A_k$ and $B_k$ at the current state for the system dynamics used in Eq. \ref{eq:dynamic_eqn} in place of the system matrices $A$ and $B$. Readers are referred to \cite{1383612}, which contains more details, including a detailed proof of the stability and feasibility of the above controller. Also, an additional disturbance can be used to compensate for the mismatch between the actual and the corresponding linearized controller assuming the non-linear model function to be a Lipschitz function \cite{gao2014}. We assume that the disturbance margin used in this work is large enough to include this extra disturbance.

\subsection{Delay-aware robust tube MPC} \label{init_state_shift}

In the system described before, we assume zero delay, i.e., we assume that the planning of the command as well as the communication of the planned command to the vehicle is instantaneous. However this is not the case in a practical vehicle system, where we have computation delay $t_c$ and control action processing delay $t_a$. Therefore, the control influences the car after $t_d = t_c + t_a$ time from the time when the observed state is used for optimization. This could lead to instability, as the robust tube assumptions no longer hold true. This is especially more dangerous when $t_d$ is large. To deal with this time delay in the system, \cite{su2013computation} proposes a bi-level control scheme where the high-level controller plans robust commands using tube MPC and the low-level unit runs a feedback controller on these reference commands. A buffer is used for communication between the two units. The brief idea is that assuming the input sequence for time frame $t$ to $t+t_h$ is known at time $t$ where $t_h$ is the horizon length, we pass the predicted $t_h$ time-ahead state obtained via simulation to the optimization problem \ref{objective function} starting at time $t$ so as to complete before $t+t_h$ time and apply the optimal input sequence to the system with feedback between times $t+t_h$ to $t+2 t_h$. The above procedure repeats in every cycle with the latest state prediction. In this strategy we need to wait for $t_h - t_c$ time for each cycle where $t_c$ is the computation time taken for optimization.     

\begin{figure}[htbp]
    \centering
    \includegraphics[width=0.5\textwidth]{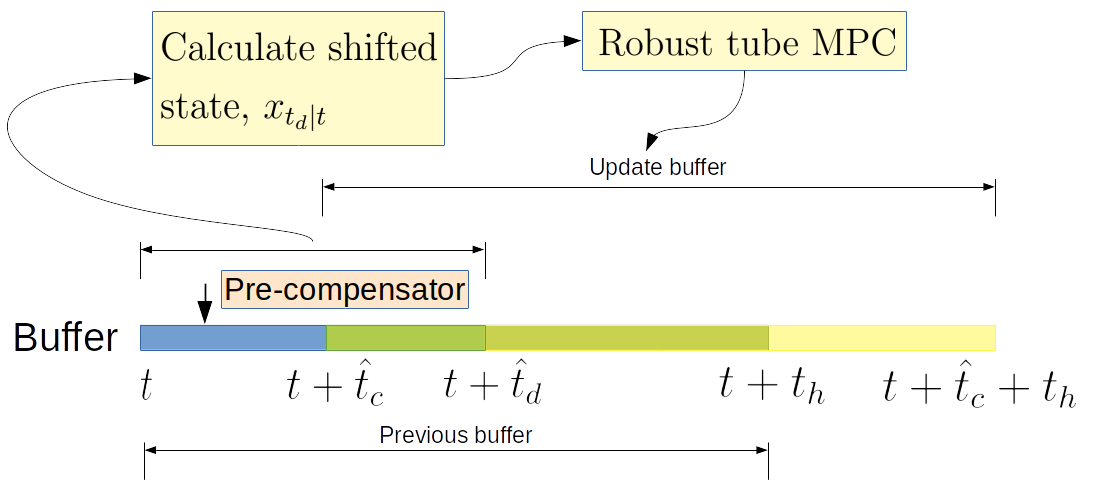}
    \caption{Our proposed delay-aware rubust tube MPC with dual cycles}
    \label{dual_cycle}
\end{figure}

However, such a common approach in the literature is inappropriate in a fast-changing environment like in the case of driving at high speeds on highways where immediate action might have to be taken in an emergency case. 
Fig. \ref{dual_cycle} illustrates our solution: we estimate a upper-bound on computation time, $\hat{t}_c$ calculated locally after each cycle with a high probabilistic guarantee. Instead of updating the buffer from $t+t_h$ to $t+2 t_h$, we update if from $t+\hat{t}_c$ to $t+\hat{t}_c+t_h$. This would significantly improve the high-level MPC controller frequency which would be required for fast-changing scenarios where the path to be followed continually updates. 
We use an adaptive Kalman filter to calculate $\hat{t}_c$. Section \ref{kalman_filter} describes this further in detail. Also, we have an additional constant time delay of $t_a$ due to control action processing, hence total upper bound on delay time is $\hat{t}_d = \hat{t}_c + t_a$. Based on this, we shift the initial state and use the shifted estimated state $\bm{x}_{\hat{t}_d|t}$ at time $\hat{t}_d$. We estimate this state at time $\hat{t}_d$ using the nominal commands from the buffer assuming no extra disturbance from the environment. We fill the buffer B from time $t + \Hat{t}_c$ to $t + \Hat{t}_c + t_h$ with the calculated controls ($\Bar{U}$) and nominal states ($\Bar{X}$). As we calculate the commands in discrete steps of $\Delta t$, the buffer is updated in the following fashion : $\Bar{\bm{u}}_{[t + \Hat{t}_c + k\Delta t, t + \Hat{t}_c + (k+1)\Delta t]} = \Bar{\bm{u}}_k$ for $k \in \{0,1,2...N-1\}$. This is demonstrated in Fig.\ref{dual_cycle}. The pre-compensator unit acts as a low-level unit which simply executes commands in the buffer at the required times and runs in parallel. It is updated more frequently than the high-level MPC.
%%%%%%%%%%%%% Eqn Here %%%%%%%%%%%%%%%

\begin{equation} \label{objective function robust}
\begin{aligned}%\label{total cost term}
    \mathop{\min}\limits_{U}&\quad\sum_{t=0}^{N-1} (\Bar{\bm{x}}_k-{\bm{x}_{ref,k}})^T Q (\Bar{\bm{x}}_k-{\bm{x}_{ref,k}})+\Bar{\bm{u}}_k^T R \Bar{\bm{u}}_k\\&\quad \quad \quad +(\Bar{\bm{x}}_N - {\bm{x}_{ref,N}})^T Q_N (\Bar{\bm{x}}_N - {\bm{x}_{ref,N}})\\
    s.t. &\quad \Bar{\bm{x}}_{k+1} = A_k \Bar{\bm{x}}_k + B_k \Bar{\bm{u}}_k\\
    &\quad \bm{x}_{t_d+t} \in \Bar{\bm{x}}_0 \oplus \mathcal{Z} \implies \bm{x}_{t_d|t} \in \Bar{\bm{x}}_0 \oplus \mathcal{Z} \ominus (\oplus_{j=0}^{s-1} A_k^j \mathcal{W})\\
    &\quad \ s = \left \lceil \frac{t_d}{\Delta t} \right  \rceil \\
    &\quad \Bar{\bm{u}} \in \mathcal{U} \ominus K\mathcal{Z}\\
    &\quad \Bar{\bm{x}} \in \mathcal{X} \ominus \mathcal{Z}
\end{aligned}
\end{equation}

\subsubsection{Control limits ($\mathcal{U}$)}

For control constraints, we limit throttle amount and steering with their actuation limits.

\subsubsection{State CBF Constraints ($\mathcal{X}$)}

\begin{figure}[htbp] 
    \centering
    \includegraphics[width=0.48\textwidth]{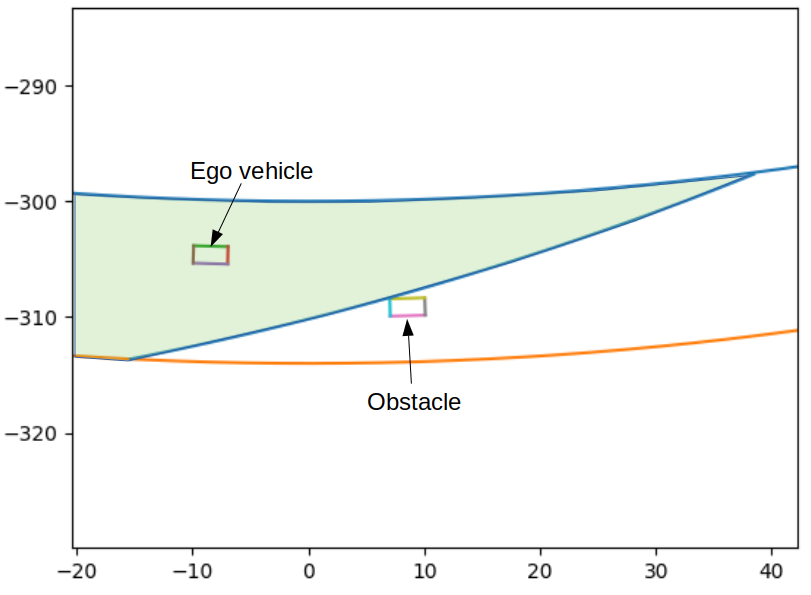}
    \caption{Computation of $X_c$ using IRIS}
    \label{iris_fig}
\end{figure}

%\qin{lateral displacement or lateral position error?}
As defined in the vehicle dynamic model of Eq. (\ref{eq:dynamic_eqn}), the longitudinal and lateral errors of the vehicle are $s$ and $e_1$. As the free space along the Frenet frame is non-convex in general, we can not directly set the constraints. This is because it becomes computationally expensive for the optimization problem to set the constraints in the non-convex form. To solve this, we propose the use of the IRIS algorithm to derive a set of convex constraints for the optimization problem to make it efficiently solvable while also ensuring safety for the vehicle \cite{khaitan2021safe}. IRIS works by finding a largest possible ellipsoid that fits the non-convex space, which is later converted to a set of convex constraints. Fig. \ref{iris_fig} shows an example of convex resultant space obtained after applying IRIS.
The lane boundaries are also fed as constraints to the non-convex free space by sampling points around the vehicle's location. The convex space $\mathcal{X}$ can be expressed as a set of linear state constraints where each bounding line can be expressed by $a_i$, $b_i$, and $c_i$ (see Eq. (\ref{cbf_constraints_A})). Further, we formulate each of these constraints with a CBF as given in Eq. (\ref{cbf_constraints_A}), where $L$ is the number of sides of the bounding polygon for the convex space $\mathcal{X}$.
% \qin{I don't understand this equation}

% Eqn Here
\begin{equation} \label{cbf_constraints_A}
\begin{split}
    &h_i(X_k) = a_i s_k + b_i e_{1,k} + c_i,  \forall i = {0,1...L-1} 
    %& a_i s_k + b_i e_{1,k} + c_i + \lambda (a_i s_k + b_i e_{1,k} + c_i) \ge 0 
\end{split}
\end{equation}
% Lane constraints can be expressed as follows by putting constraint on the lateral displacement, $e_1$ assuming constant lane width, $W$.  
% An example of convex constraints after applying IRIS is shown in Figure \ref{iris_fig}.
%%%%%%%%%%%%%%%%%%% Fig here %%%%%%%%%%%%%%%%%%%%%% 

\subsubsection{Disturbance-invariant set ($\mathcal{Z}$)}

The disturbance-invariant set is the set within which the system is guaranteed to be contained given the bounded disturbance set, $W$. We can over-approximate $Z$ for the $k^{th}$ time step as : $\mathcal{Z}_k = \Sigma^{N}_{i=0} A_{k}^{i} \mathcal{W}$ \cite{article}. However for a LTV system with delay, the linearized control matrix $A_{k+1}$ is dependent on the current state, $x_k$, hence it would be different for each control cycle. So to ensure robustness, $\mathcal{Z}_{k}$ must be covered by $\mathcal{Z}_{k+1}$. For a detailed proof on this, readers are referred to our previous work \cite{dvij2022}.

\subsection{Estimating computation time} \label{kalman_filter}
\begin{figure}[htbp] 
    \centering
    \includegraphics[width=0.5\textwidth]{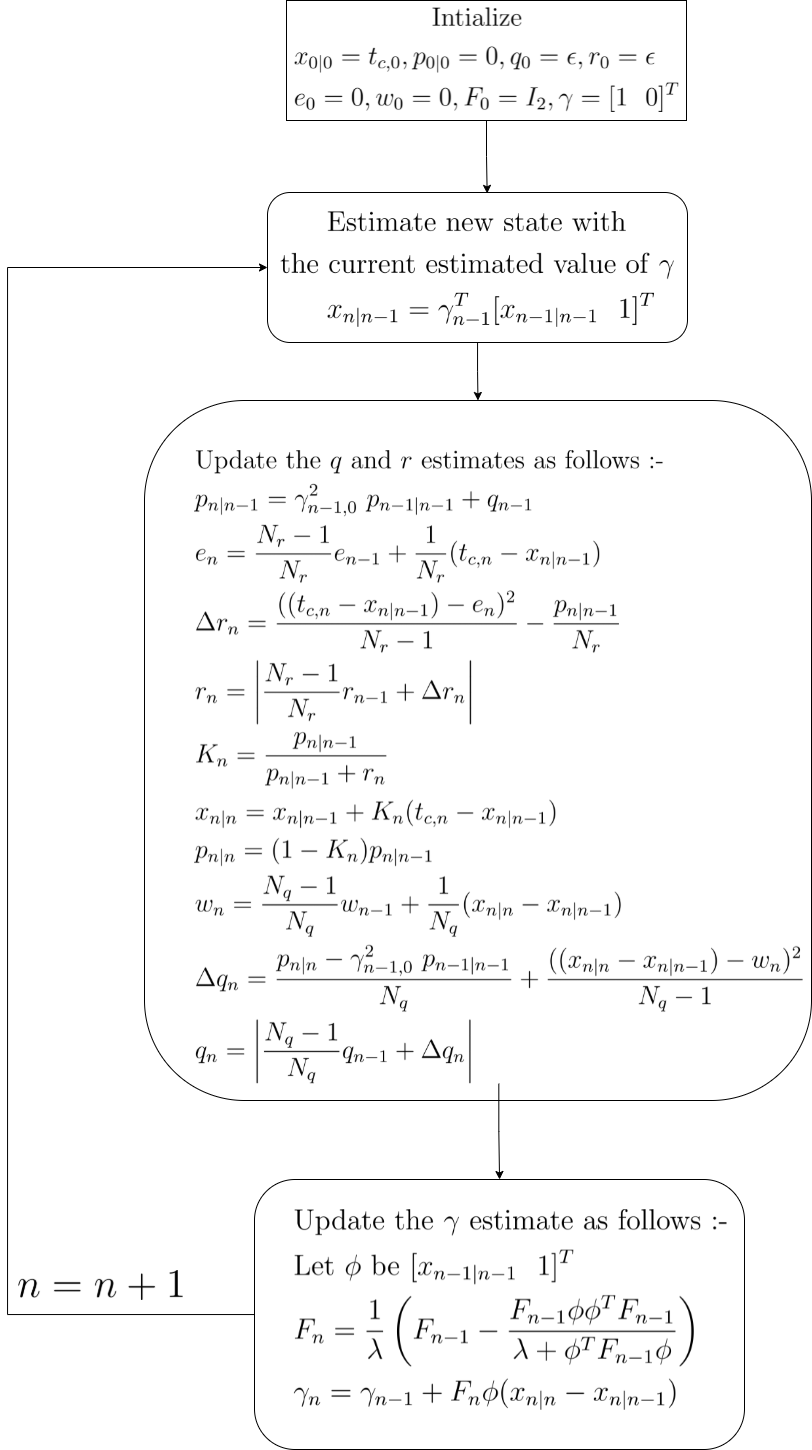}
    \caption{INFLUENCE algorithm}
    \label{influence_fig}
\end{figure}

We propose the use of INFLUENCE for the estimation of an upper bound on computation time. It is a variant of the adaptive Kalman filter. The conventional Kalman filter is unsuitable for cases in which the process model and the noises covariance are unknown. These are compensated by using adaptive filters. However, state-of-the-art adaptive filters rely either on known dynamics while assuming unknown noise covariances \cite{article_kalman,myers1976adaptive} or on known covariances while assuming unknown model parameters \cite{liu2015safe}. To compensate for this deficiency, INFLUENCE assumes that measurement noise variance $r$, process noise variance $q$ as well as $\gamma$ in the process model,  are all unknown. We assume a linear unknown process model with parameters $\gamma$. We assume both process and observation noise distributions to be mutually uncorrelated, independent and Gaussian. INFLUENCE works as described in Fig. \ref{influence_fig}, where $t_{c,n}$ is the observed computation time, and $x_{n|n-1}$, $p_{n|n-1}$ are the predicted computation time mean and variance estimates for the $n^{th}$ time step.

The flowchart starts with initialization of all variables. The key idea is that we use the model information $\gamma$ in the current step to estimate the noise information $q$ and $r$, followed by the update of $\gamma$ itself \cite{article_kalman,myers1976adaptive}. In the third box, $w$ and $e$ are measurement and prediction errors. They are updated in an incremental manner. $N_q$ and $N_r$ are tunable parameters for update rates. In the last box, $\lambda$ and $F$ can be considered as a forgetting factor and a learning rate, respectively.

It is important to note here that for the sake of simplicity, the notation $x$ is used for the computation time to be estimated and is not to be confused with the state notation we used earlier. We use the predicted mean $x_{n|n-1}$ and variance $p_{n|n-1}$ estimates for the $n^{th}$ time step to get an upper-bound estimate on computation time: $\Hat{t}_{c,n} = x_{n|n-1} + \beta p_{n|n-1}$. Assuming a Gaussian distribution, the parameter $\beta$ dictates the level of confidence in estimating $\Hat{t}_{c,n}$ as an upper bound (higher value of $\beta$ implies higher confidence). For the experiments in this paper, we use $\beta = 2$.

\subsection{Controller Plan A}

\begin{figure}[htbp] 
    \centering
    \includegraphics[width=0.4\textwidth]{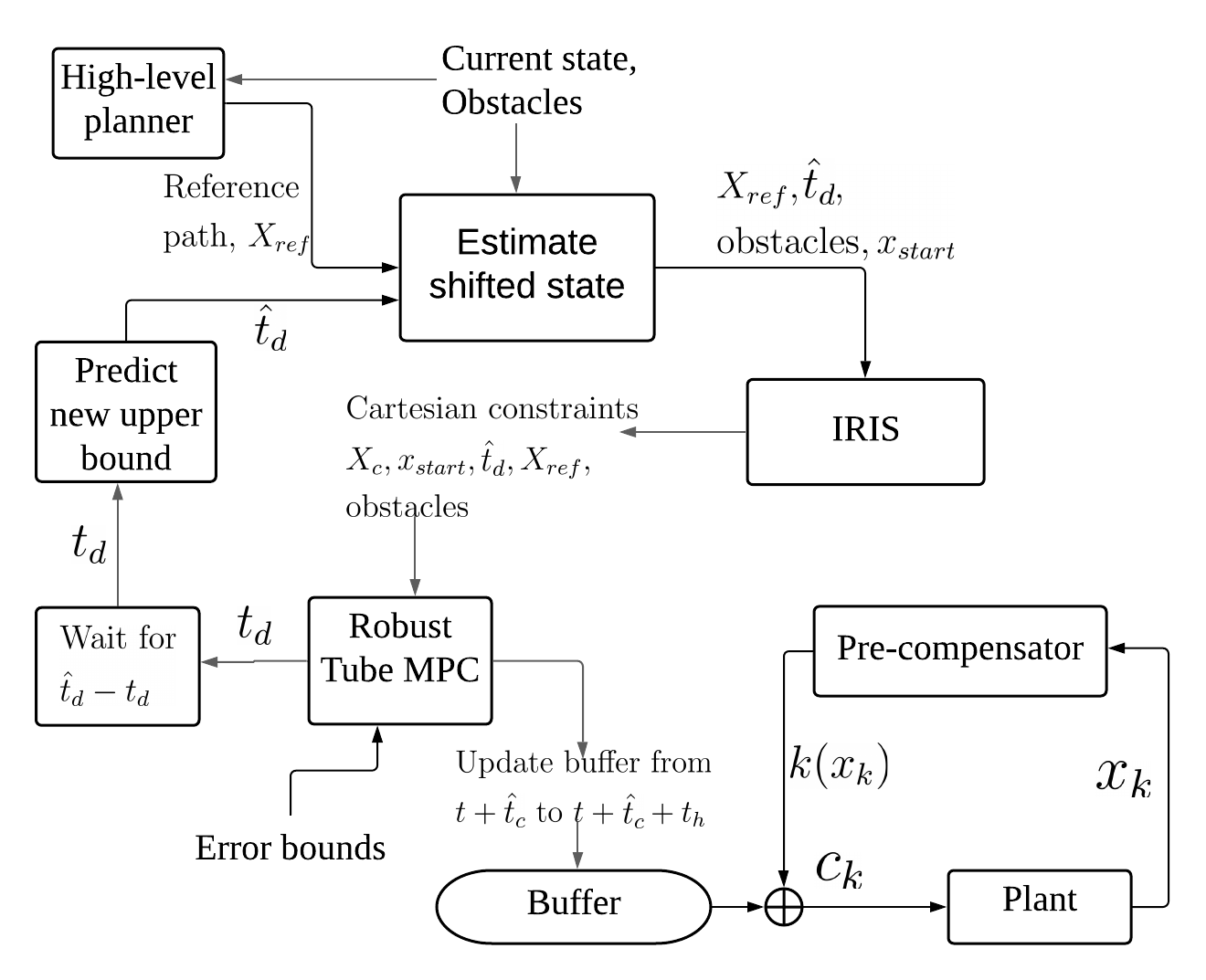}
    \caption{Controller Plan A for delay-aware robust tube MPC.}
    \label{planA_fig}
\end{figure}

Putting everything together, we propose the first controller design as illustrated in Fig. \ref{planA_fig}. The steering dynamic delay is compensated by adding the actual steering state as a new state and by using a first-order ODE to model steering dynamics (see section \ref{actuator_dynamics}). The computation and actuator dynamic delays are handled by initial state shifting (see \ref{init_state_shift} and \ref{kalman_filter}). After the optimization, the calculated commands from robust tube MPC are used to update the buffer from $t+\hat{t}_c$ to $t+\hat{t}_c+t_h$ filled with the nominal states and the commands. Observed computation time is used to update the new local upper bound for the next cycle by using our filter. The pre-compensator unit runs in parallel with high frequency as a feedback controller using the nominal states and commands from the buffer.        
% We use initial state shift by the estimated local upper bound calculated using INFLUENCE (Section \ref{kalman_filter}) to compensate for the computation and actuator command processing delay. After the full cycle, the calculated commands from robust tube MPC are used to update the buffer from $t+\hat{t}_c$ to $t+\hat{t}_c+t_h$ with the nominal commands and states. In parallel, the pre-compensator unit runs as a low-level process with a high frequency to execute the refined controller commands.
% The first controller design (called plan A) is illustrated in Fig. \ref{planA_fig}. We compensate for the actuator's steering dynamics by modelling a first-order ODE. For the compensation of the computation and actuator command processing delays, we use initial state shift by the estimated local upper bound on the net delay time. The optimization problem updates the robust tube buffer from $t+\hat{t}_c$ to $t+\hat{t}_c+t_h$ with the nominal commands and states. The pre-compensator unit runs as a low-level process to refine the control with a higher frequency (see Fig. \ref{planA_fig}).

%%%%%%%%%%%%%%%%%%% Fig Here %%%%%%%%%%%%%%%%%%%%%%%
% If the true delay $t_d$ is smaller than the estimated $\hat{t}_d$, we need to wait $\hat{t}_d-t_d$ for a time alignment.
\subsection{Controller Plan B}

\begin{figure}[htbp] 
    \centering
    \includegraphics[width=0.4\textwidth]{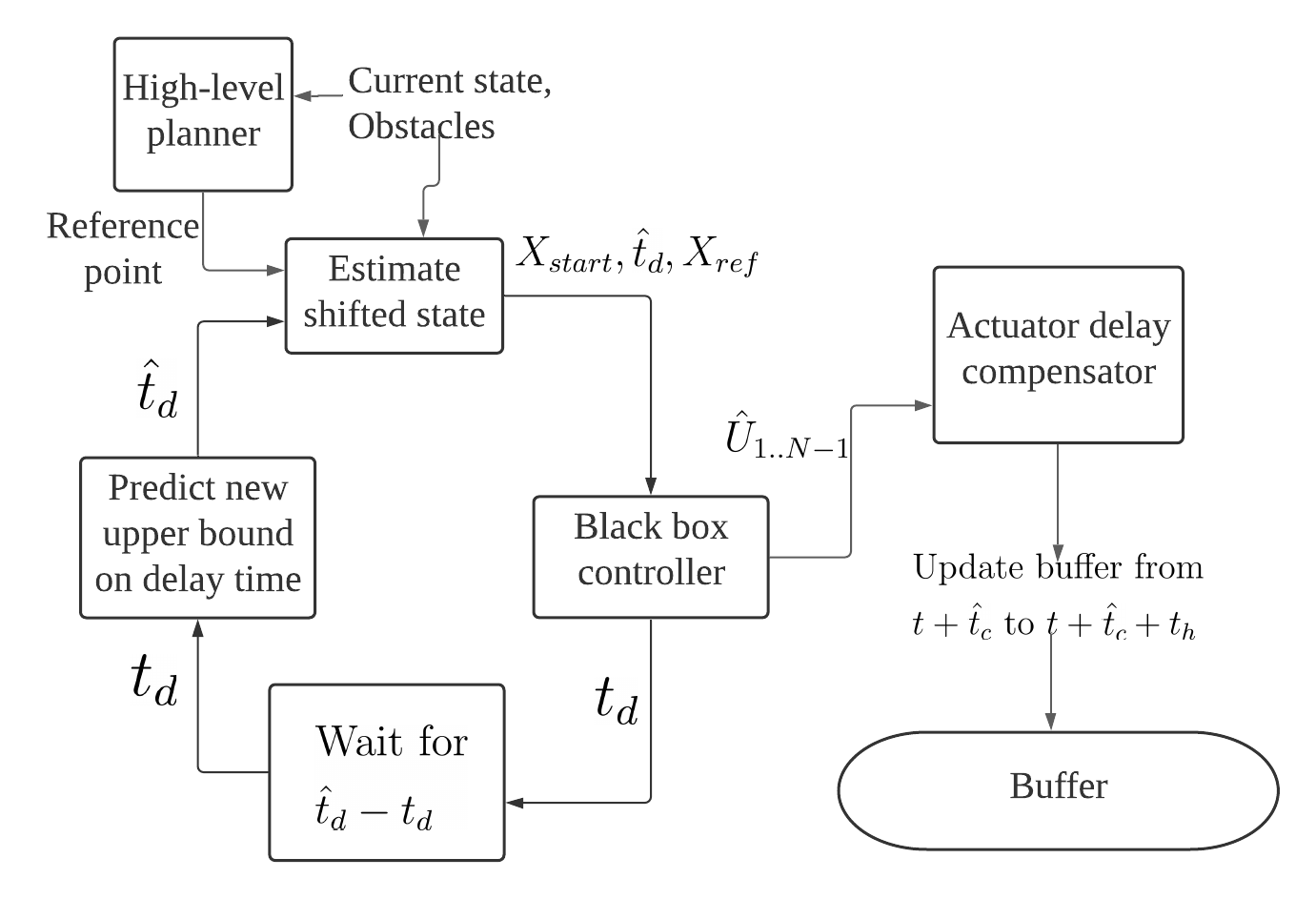}
    \caption{Controller Plan B for black box controller.}
    \label{planB_fig}
\end{figure}

% \qin{We need a better expression here, because I use u for the control vector, but here only the subspace, the steering, is used, please discuss it with me}
We also propose a new controller plan aimed towards safeguarding blackbox controllers whose internal mechanism is not known, like a Learning Enabled (LE) controller which has been trained in a simplistic simulation environment where practical delays are not considered. We call the LE controller the nominal controller (see Fig. \ref{planB_fig}). To compensate for the computation time and control action processing delay, we use a similar design to that of plan A through initial state shift. However, for compensating actuator dynamic delay, we propose the use of a separate process which uses QP to output control commands that closely track the desired nominal commands inputted to it while also avoiding collision and respecting lane constraints. We describe this formulation in detail later. First, we estimate the computation and control action processing delay and use it to shift the initial state, which is inputted to the LE controller. Rollouts are conducted to obtain sequential commands as output from the LE controller expressed as $\Hat{U} = \{\Hat{\bm{u}}_1,\Hat{\bm{u}}_2,\cdots,\Hat{\bm{u}}_N\}$. After solving the QP problem in Eq. (\ref{actuator_dynamics_compensator}) we get the refined commands as output expressed as $U = \{\bm{u}_1, \bm{u}_2, \cdots, \bm{u}_N\}$. Here, $u_{start}$ is the observed steering angle at the start before optimization and $r_k$ is obtained by taking output at time $k \Delta t$ in response to a unit step input to the steering actuator at time $0$. $Q_{ac}$ and $R_{ac}$ are positive semidefinite weight matrices for tracking controller commands $\Hat{U}$ and minimizing controller effort, respectively. Along with actuator dynamic delay compensation, we also add CBFs to offer safety constraints to the blackbox controller.

\subsubsection{Lane constraints} \label{subsec:lane_constraints}

We assume variable left and right lane width as a function of the longitudinal displacement $s$ along the racing line. Let the approximate linear functions $L_l(s)$ and $L_r(s)$ govern the left and right lane width functions, respectively. Based on this, we have
% We set the CBF (see (Eq. \ref{lane_cbf_B})) for enforcing lane constraints at each step. We take into account the variations in the left and right lane widths along the reference racing line by approximating the lane width as a linear function of the longitudinal displacement $s$. Let $L_l(s)$ and $L_r(s)$ be the lane widths on the left and the right of the reference racing line.

% Eqn here
\begin{equation} \label{lane_cbf_B}
\begin{split}
    &h_{left,k}(\bm{x}) = L_l(s_0) - e_{1,k} \\
    &h_{right,k}(\bm{x}) = e_{1,k} + L_r(s_0)\\
    &-L_r(s_0) \le \frac{1}{\lambda^2} (\ddot{e}_{1,k} + \lambda \dot{e}_{1,k} + \lambda^2 e_{1,k}) \le L_l(s_0)\\
    &
\end{split}
\end{equation}

\subsubsection{Collision avoidance constraints}
For collision avoidance with each vehicle, we set the following second-order CBF since the dynamic model has a relative degree of two. The detailed justification and expansion can be found in the Appendix.

\begin{equation}
    h_{veh,k}(\bm{x}, \bm{x}_{opp}) = \left(\frac{s-s_{opp}}{d_s}\right)^2 + \left(\frac{e_1 - e_{opp}}{d_e}\right)^2-1
\end{equation}
where $s_{opp}$ and $e_{opp}$ are the longitudinal and lateral error of the opponent vehicle, and $d_s$ and $d_e$ are the longitudinal and lateral safe distances from the target opponent vehicle. $\bm{x}_{opp}$ is the opponent vehicle's state. The safety constraint of this second-order CBF is: 
\begin{equation}
    h_{veh,k}(\bm{x}, \bm{x}_{opp}) + 2 \lambda \dot{h}_{veh,k}(\bm{x}, \bm{x}_{opp}) + \lambda^2 \ddot{h}_{veh,k} \ge 0
\label{Eqn:collision_constraints}
\end{equation}

\begin{equation} \label{actuator_dynamics_compensator}
    \begin{split}
    \mathop{\min}\limits_{U}&\quad\sum_{k=1}^{N} \left\Vert \Hat{\delta}_k - (\delta_0 + \sum_{i=1}^{i=k} (\delta_i - \delta_{i-1}) r_{k-i+1}) \right\Vert_{Q} + \left\Vert \delta_k \right\Vert_{R} \\
    s.t. &\quad \bm{u}_0 = \bm{u}_{start}\\
    &\bm{x}_{k+1} = A_k \bm{x}_k + B_k \bm{u}_k\\
    &\quad h(\bar{\bm{x}}_k, \bm{x}_{opp}) + \lambda \dot{h}(\bar{\bm{x}}_k, \bm{x}_{opp}) + \lambda^2 h(\bar{\bm{x}}_k, \bm{x}_{opp}) \ge 0
    \end{split}
\end{equation}
where $r_i = (1 - e^{-K i \Delta t})$.

\section{Experimental Results}
\label{sec:experiment}
In this section, we describe the experimental setups, as well as report and discuss all of the tests conducted in the Gazebo \cite{koenig2004design} and the CARLA \cite{dosovitskiy2017carla} simulators. The Gazebo simulator is used to test normal autonomous driving scenarios (which we refer to as non-racing in the rest of the paper). The racing scenarios are tested using the CARLA simulator. Validation of controller plan A and controller plan B is divided into two subsections.

First of all, to deal with the modeling of steering delay, we need to collect steering data from the vehicle and identify the delay parameter. The basic idea is to test the unit step response. The detailed process can be found in \cite{dvij2022}. %We demonstrate the process in the Gazebo simulator. The identification in CARLA is similar. We use a Prius vehicle model in the Gazebo. In order to get the time constant value for the steering actuator, we test its step response, i.e., we set the actuator command to a constant reference and record the steering angle values over a time window sufficient for the steering angle to converge to the maximum value. We then fit the observed response values with the first-order ODE described in Section \ref{actuator_dynamics} and determine the parameter $K_{\delta}$. As shown in Fig. \ref{response_fig}, using $K_{\delta} = 30$ well approximates the actuator dynamics for the Prius model in Gazebo.

%%%%%%%%%%%%%%%%%%% Fig Here %%%%%%%%%%%%%%%%%%%%%%%
% \begin{figure}[htbp]
%     \centering
%     \includegraphics[width=0.35\textwidth]{responses.png}
%     \caption{Steering angle response for Prius model in Gazebo}
%     \label{response_fig}
% \end{figure}

\subsection{Controller plan A}

\subsubsection{Static obstacle avoidance (non-racing)} \label{sec:exp1A}

\begin{figure}[htbp] 
    \centering
    \includegraphics[width=0.45\textwidth]{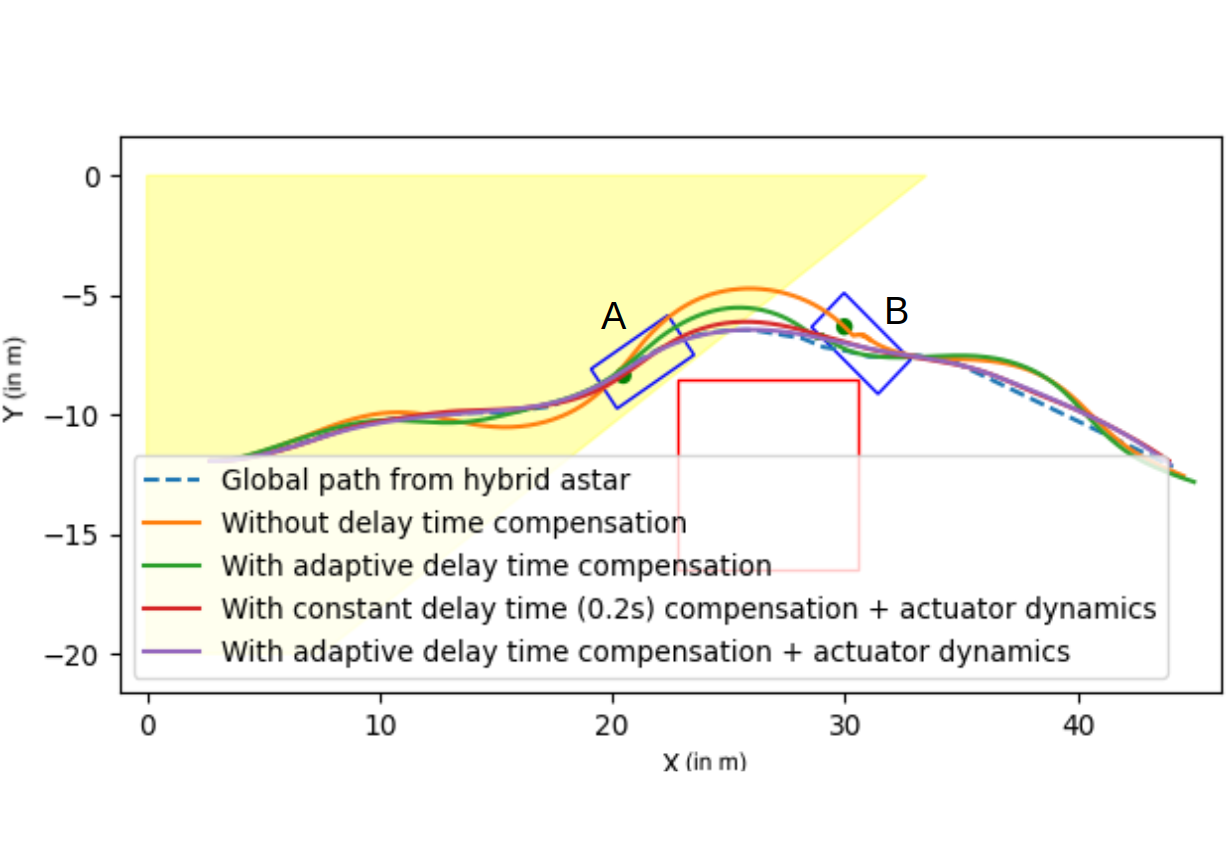}
    \caption{Path following comparison in the experiment \ref{sec:exp1A}. Yellow zone is pose A's corresponding convex constraint computed by IRIS. Red box: static obstacle.}
    \label{fig:fig1_paths}
\end{figure}

\begin{figure}[htbp]
    \centering
    \includegraphics[width=0.4\textwidth]{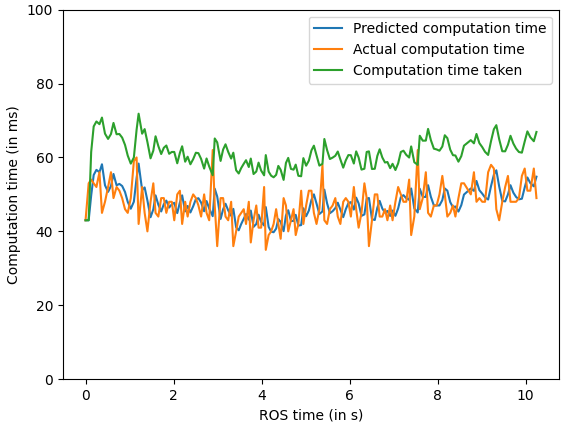}
    \caption{Experiment \ref{sec:exp1A}, predicted computation time (blue): predicted value using INFLUENCE without adding variance; computation time taken (green): predicted value using INFLUENCE with variance; actual computation time (orange): ground truth delay.}
    \label{fig:fig1_times}
\end{figure}

We conduct a static obstacle avoidance experiment to test controller plan A. In the experiment  shown in Fig. \ref{fig:fig1_paths}, we have the following observations: 1) without considering delays (orange trajectory), the vehicle has a larger overshoot. A collision happens at pose B. 2) our controller (purple trajectory) has smoother tracking than the fixed delay approach (red trajectory) and the controller only considering computation delay (green trajectory). As shown in Fig. \ref{fig:fig1_times}, INFLUENCE can safely bound the computation delay in this experiment.

% We compare the controller's path following with and without considering delays. In the absence of any compensation, the vehicle overruns the reference line during the first run, resulting in higher tracking error, and when it attempts to return safely to the reference line, it collides with the static obstacle, as indicated by pose B in Fig \ref{fig:fig1_paths}. When delay compensations are considered, the path followed by the vehicle is closer to the reference line with less tracking error and is collision-free. We also make a comparison with the following two cases:  1) when the delay time is taken as the upper bound equal to the horizon length \cite{su2013computation} and 2) using INFLUENCE to find the local upper bound estimate. As observed in Fig \ref{fig:fig1_paths}, the path followed in the second case with variable delay time is clearly smoother and has less tracking error compared to the first case. This is because in the first case, where constant delay compensation is considered, during the first turn at pose A, the state constraints generated by IRIS force the vehicle to move away from the reference line to ensure safety, as shown in Fig \ref{fig:fig1_paths}, but the vehicle is able to return safely to the reference, since delay compensation is considered. On the other hand, in the second case where we use INFLUENCE to estimate the local upper bound on delay time, due to relatively less response time, the controller responds faster to changing IRIS constraints after moving past the first turn, and hence has relatively less overshoot.


\subsubsection{Racing scenario} \label{sec:exp3A}
\begin{figure}[htbp]
    \centering
    \includegraphics[width=0.5\textwidth]{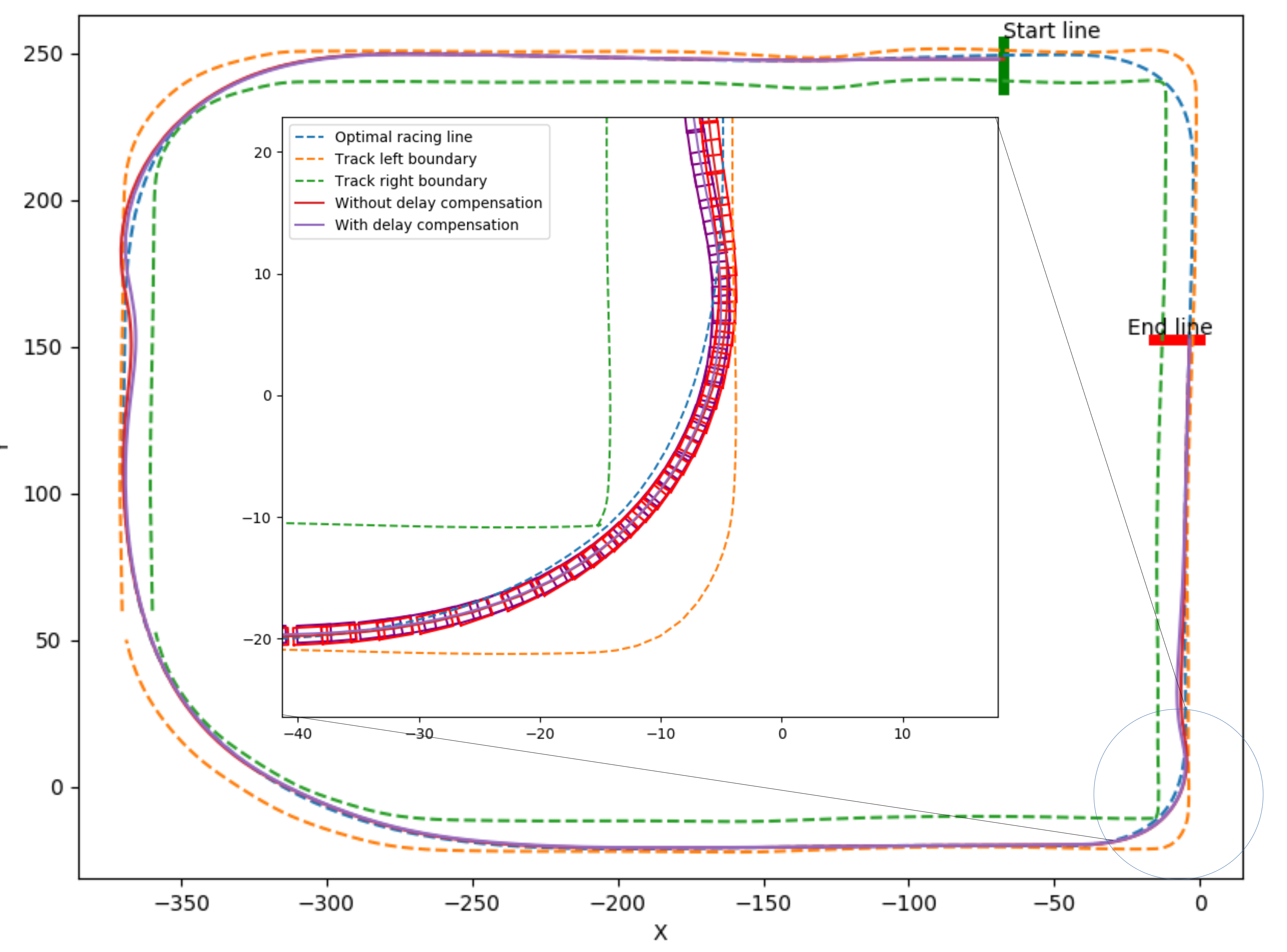}
    \caption{Comparison in paths followed with and without delay compensation by racing car. The vehicle moves counterclockwise}
    \label{fig:fig_track_exp3A}
\end{figure}

We set up a racing scenario in CARLA to validate the controller plan A. A part of the track from the Town06 map is used for the experiments. It is a closed track suitable for setting up a racing scenario. The optimal racing line for the track is given in Fig. \ref{fig:fig_track_exp3A}. In the case when no delay compensation is considered (see Fig. \ref{fig:fig_track_exp3A}), the vehicle crashes once during the lap at a sharp turn. The reason for this is that as the vehicle operates at its friction limit at the turn, tracking error accumulates due to computation, and actuator delays lead the vehicle to hit the boundary (see the zoomed-in plot of the box (control without delay compensation) overlap with road boundary), thus leading to high lap time. On the other hand, with the delay compensation plan included (see Fig. \ref{fig:fig_track_exp3A}, the vehicle safely and closely tracks the racing line, leading to faster lap time. We further compare race times and the number of crashes with the lane boundary (see Tab. \ref{table:race_times}). As can be observed, with delay compensation, the vehicle takes less time to complete the race, while without delay compensation due to crash, the vehicle takes more time to complete the race.
% \qin{difficult to distinguish red line and purple line, discuss with me})
% \qin{missing in the figure?}

% \begin{figure}[htbp]
%     \centering
%     \includegraphics[width=0.5\textwidth]{Exp3A/speeds_without.png}
%     \caption{Ego vehicle speed for the case when delay compensation is not considered \qin{can be deleted due to tiny difference}}
%     \label{fig:fig_speeds_without_exp3A}
% \end{figure}

% \begin{figure}[htbp]
%     \centering
%     \includegraphics[width=0.5\textwidth]{Exp3A/speeds_with.png}
%     \caption{Ego vehicle speed for the case when delay compensation is considered\qin{can be deleted due to tiny difference}}
%     \label{fig:fig_speeds_with_exp3A}
% \end{figure}

\begin{table}[htbp]
\begin{center}
\begin{tabular}{m{2.5cm} m{2cm} m{2.5cm}} 
 \toprule
 Run & Lap time (in s) & No of lane boundary violations \\ [0.5ex] 
 \midrule
 Without delay compensation & 35.46 & 1 \\ 
 With delay compensation & \textbf{35.34} & \textbf{0} \\
 \bottomrule
\end{tabular}
\end{center}
\caption{Comparison in lap times with and and without delay compensation included}
\label{table:race_times}
\end{table}

\subsubsection{Sudden change scenario (racing)} \label{sec:expD}

\begin{figure}[htbp]
\begin{subfigure}{.5\textwidth}
    \centering
    \includegraphics[width=\textwidth]{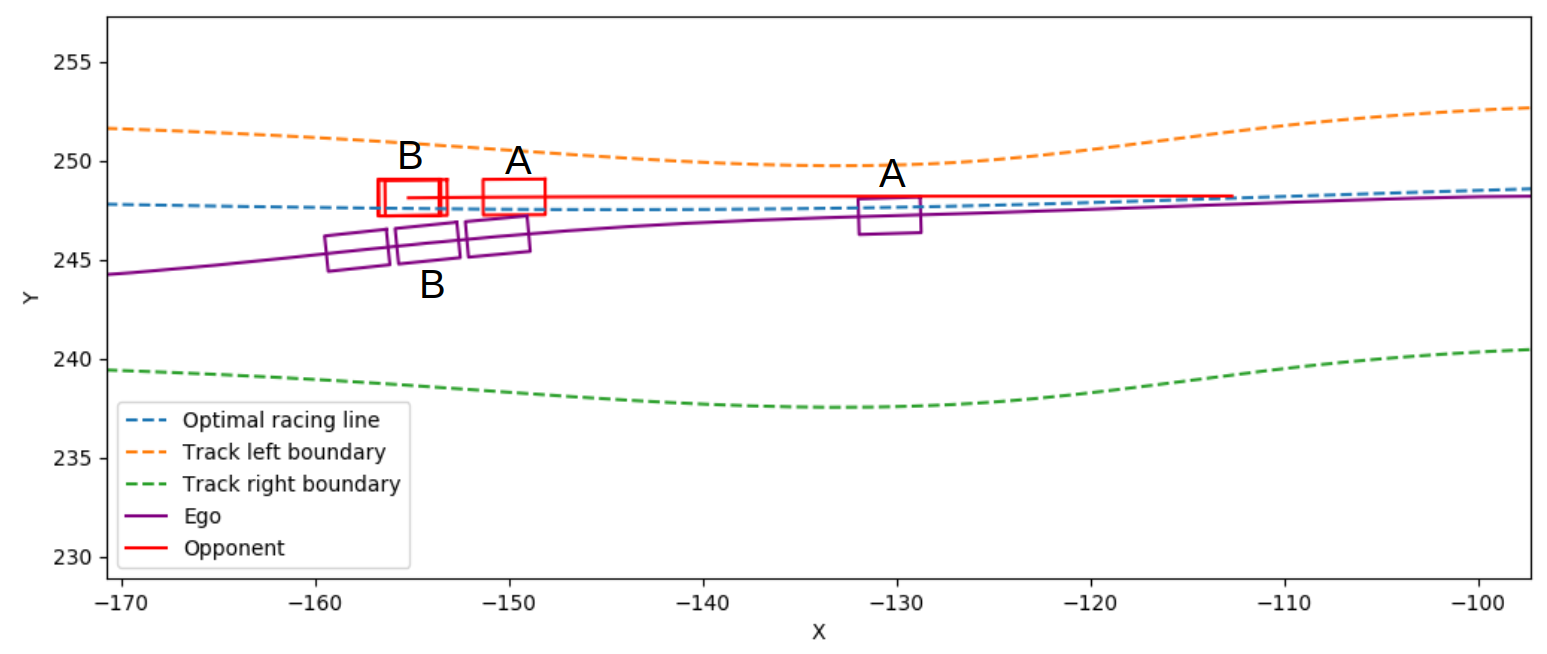}
    \caption{changing local upper bound of computation delay (ours).}
    \label{fig:4A_with_comp}
\end{subfigure}
\newline
\begin{subfigure}{.5\textwidth}
    \centering
    \includegraphics[width=\textwidth]{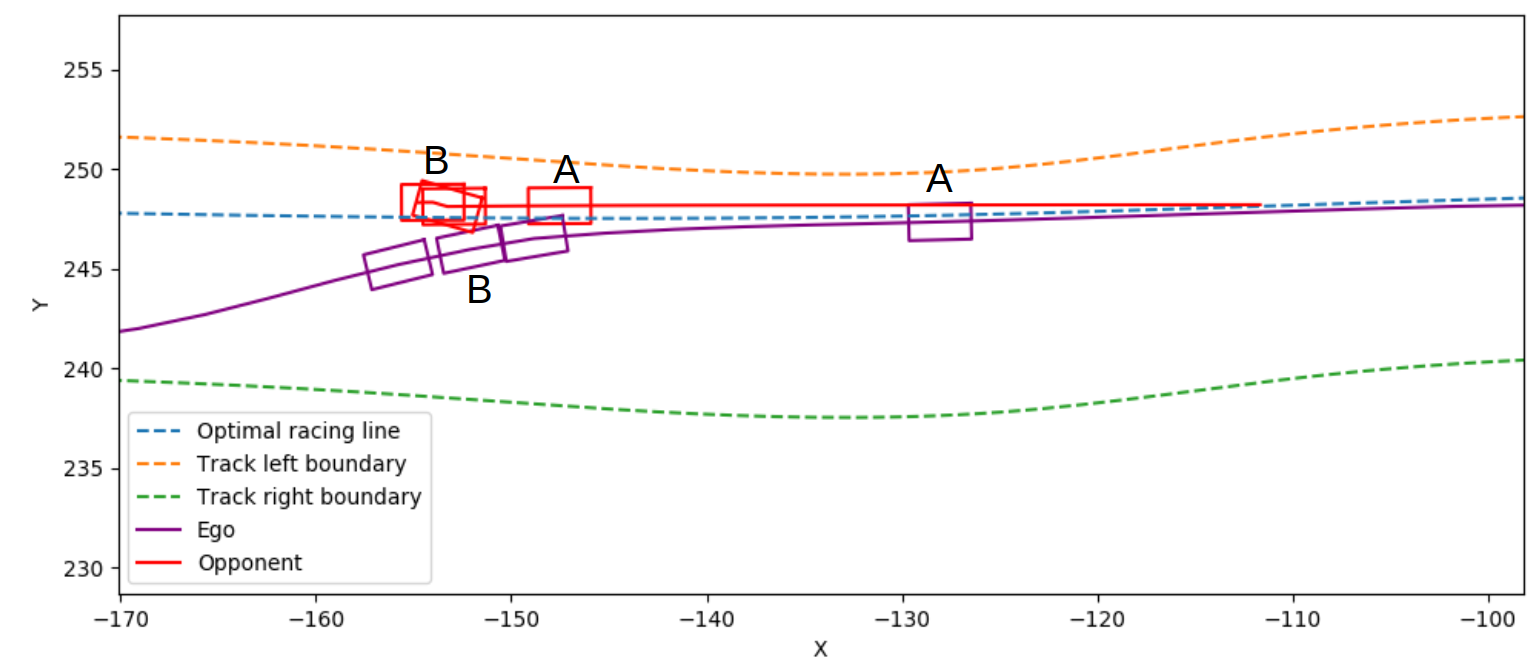}
    \caption{No computation delay.}
    \label{fig:4A_without_comp}
\end{subfigure}
\caption{Comparison in Experiment \ref{sec:expD}. The opponent vehicle starts braking at point A. The ego vehicle crashes into the opponent vehicle at point B in the case in which delay compensation is not considered. }
\end{figure}

We also test controller plan A against a sudden change racing scenario. For this scenario, an opponent vehicle ahead of the ego vehicle suddenly loses control, and thus applies sudden braking at $t=2s$, forcing the ego vehicle to respond suddenly to avoid collision. If delay compensation is not considered for this scenario (see Fig. \ref{fig:4A_without_comp}), the vehicle collides with the opponent vehicle. On the other hand, with delay compensation considered, the vehicle makes a more informed decision to make a sudden turn with application of brakes (see Fig. \ref{fig:4A_with_comp}) to avoid collision with the opponent vehicle.

\subsubsection{Overtaking in a turn scenario (racing)} \label{sec:exp5A}

\begin{figure}[htbp]
\begin{subfigure}{.5\textwidth}
    \centering
    \includegraphics[width=\textwidth]{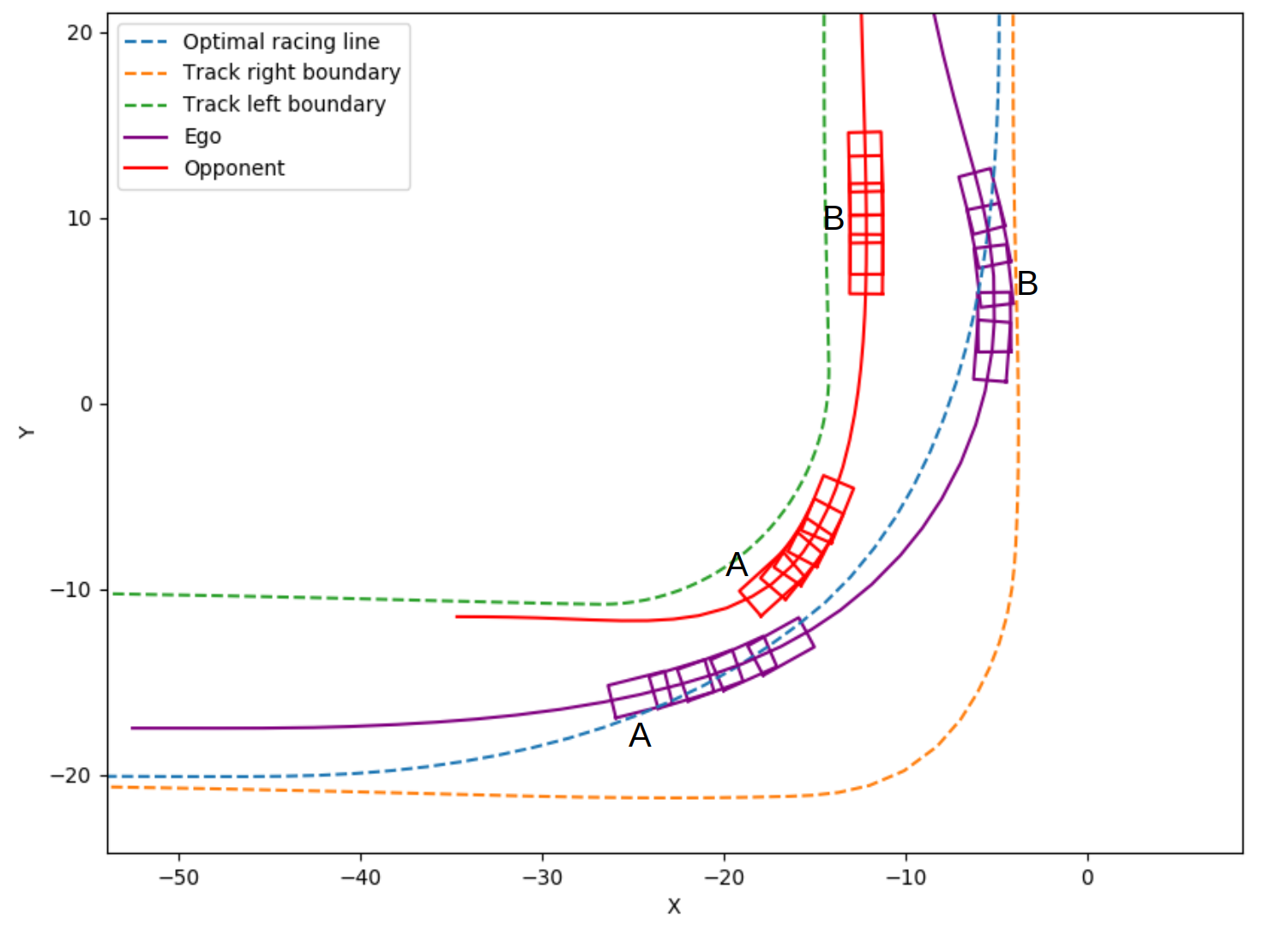}
    \caption{With delay compensation.}
    \label{fig:5A_with_comp}
\end{subfigure}
\newline
\begin{subfigure}{.5\textwidth}
    \centering
    \includegraphics[width=\textwidth]{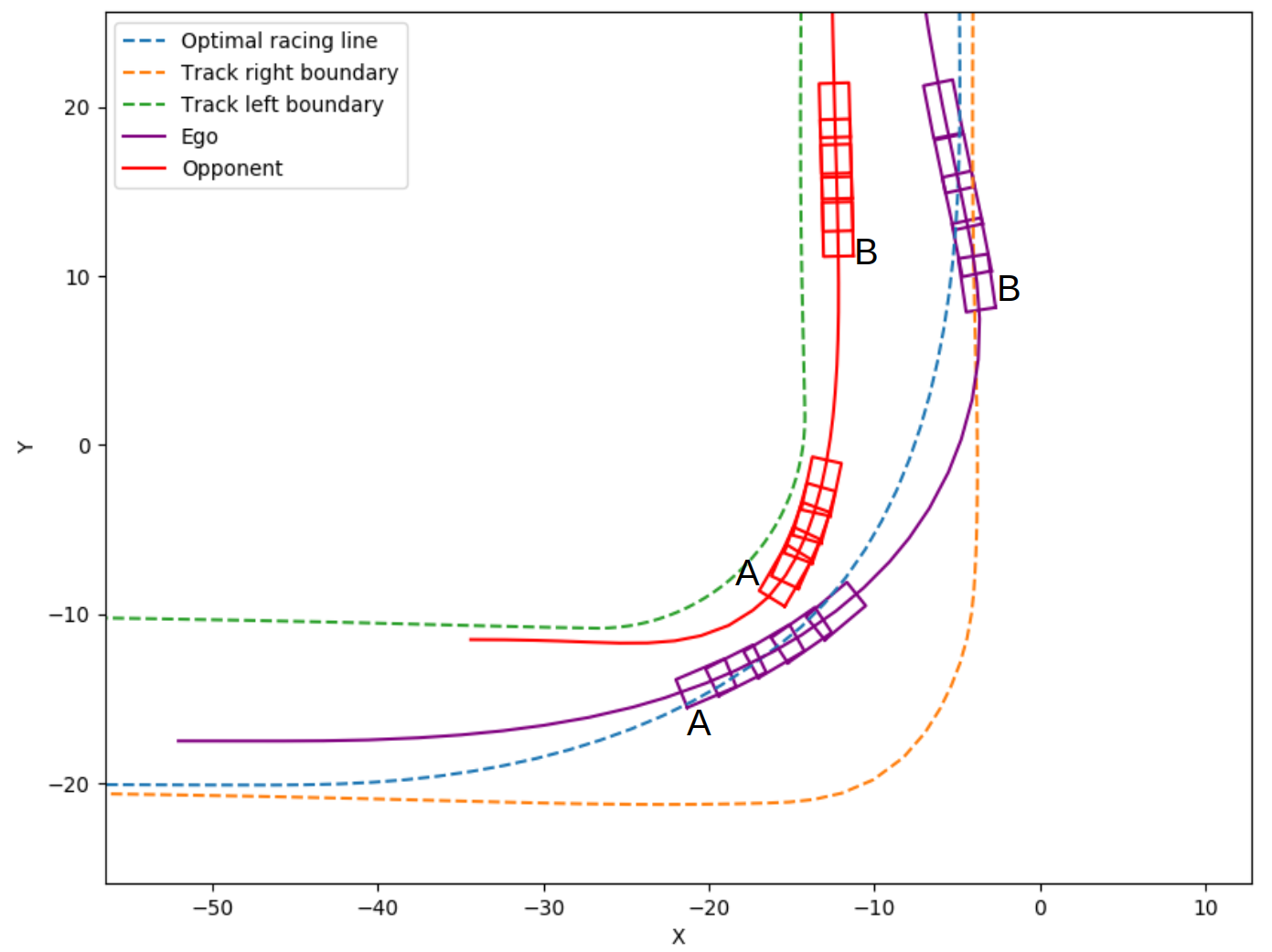}
    \caption{Without delay compensation.}
    \label{fig:5A_without_comp}
\end{subfigure}
\caption{Comparison in Experiment \ref{sec:exp5A}. The ego vehicle crashes at point A in the case in which delay compensation is not considered. }
\end{figure}

Finally, we test the controller plan A in a scenario where the vehicle has to overtake at a turn, forcing it to leave the optimal racing line at the turn. Such a scenario requires highly precise controls, as the vehicle operates at its friction limits. As the vehicle deviates from the optimal racing line to overtake and avoid collision, the vehicle has to apply braking to avoid moving out of the track. However, without the delay compensation (see Fig. \ref{fig:5A_without_comp}), the vehicle applies braking late, leading to the vehicle CBF constraint being unable to retain the vehicle within the lane boundaries. With delay compensation considered, the vehicle makes more effective control decisions taking the delay into account and thus is able to perform the overtaking maneuver safely at turns (see Fig. \ref{fig:5A_with_comp}).

\subsection{Controller plan B}
We test the controller plan B in two scenarios described below. Due to convenience of training, imitation learning is used in our work for training a blackbox LE controller, which has a demonstrator (a model predictive controller) to obtain state-to-control mapping data \cite{claviere2019trajectory}. We use a fully-connected feed-forward neural network (FNN) as a LE controller. The training dataset is from samples of the demonstrator, i.e.,  $\mathcal{D} = {(\bm{\xi}_0, \bm{u}_0),(\bm{\xi}_1, \bm{u}_1), \cdots, (\bm{\xi}_n, \bm{u}_{n})}$, where $\bm{\xi_i} = [\delta x, \delta y, \delta \theta, v_{ego}, v_{target}]^T$, with ($\delta x, \delta y$) and $\delta \theta$ respectively the relative position and orientation of the target waypoint, and $v_{ego}$ and $v_{target}$ respectively the current ego vehicle speed and target speed at the waypoint. For the racing scenario, the optimal racing line is used as the reference for tracking with the target speed, $v_{target}$, as the reference speed at the target waypoint. The output $\bm{u}_i= [\delta_i, p_i]$, where $\delta$ is the steering angle and $p$ is the pedal. The cost function can be expressed as a squared loss function: $J = \sum_{i=0}^{n}||F(\bm{\xi}_i)-\bm{u}_i||_2$. The training is done by using back-propagation to minimize the loss offline.

\subsubsection{Path following in racing scenario} \label{sec:exp1B}

\begin{figure}[htbp]
    \centering
    \includegraphics[width=0.5\textwidth]{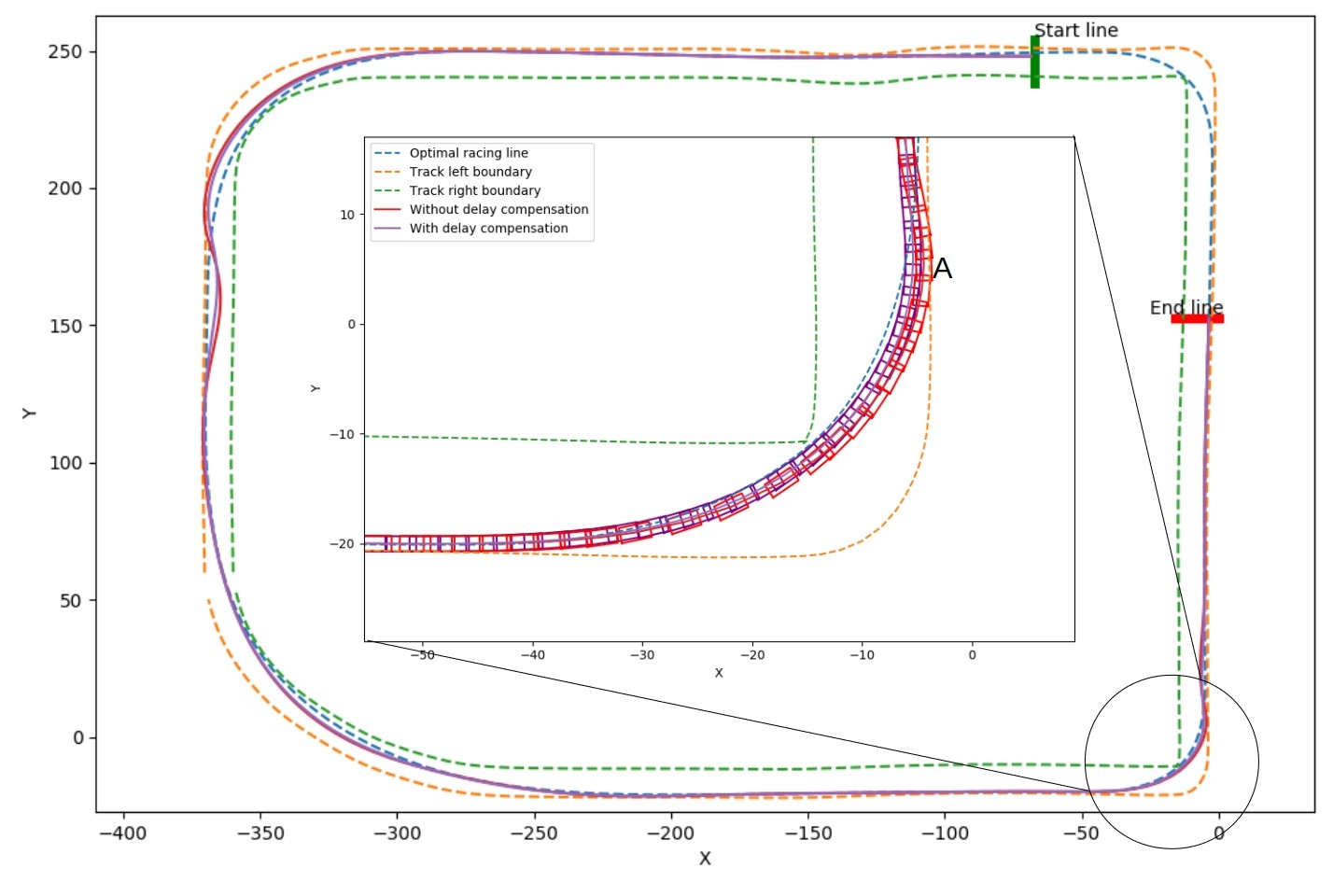}
    \caption{Comparison in paths followed with and without delay compensation by racing car. The vehicle moves clockwise. }
    \label{fig:fig_track_exp1B}
\end{figure}

We test controller plan B in the path following racing scenario without obstacles. As discussed, as the track requires highly precise controls at turns to stay within the track, the safety controller in the form of lane constraints (see the CBF formulation in Sec. \ref{subsec:lane_constraints}) has to come into play so as to compensate for errors in current state measurements and the delays. As shown in Fig. \ref{fig:fig_track_exp1B}, when delay compensation is not considered the vehicle moves out of the racing line as the safety constraints are not able to safeguard the vehicle due to inconsistency in vehicle state caused by delay errors. However, with delay compensation (see Fig. \ref{fig:fig_track_exp1B} and Fig. \ref{fig:fig_speeds_with_exp1B}), the vehicle is able to safely stay within the lane constraints.

\begin{figure}[htbp]
    \centering
    \includegraphics[width=0.5\textwidth]{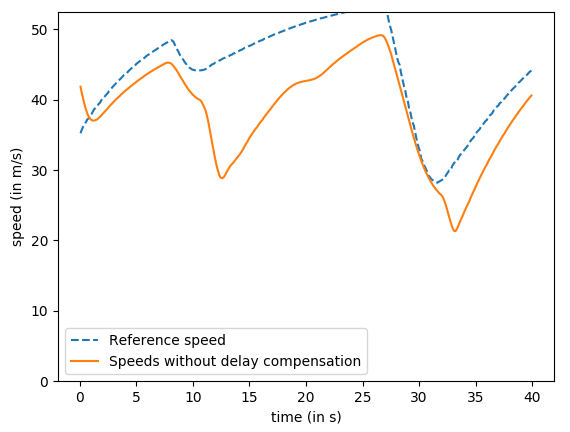}
    \caption{Ego vehicle speed for the case when delay compensation is not considered}
    \label{fig:fig_speeds_without_exp1B}
\end{figure}

\begin{figure}[htbp]
    \centering
    \includegraphics[width=0.5\textwidth]{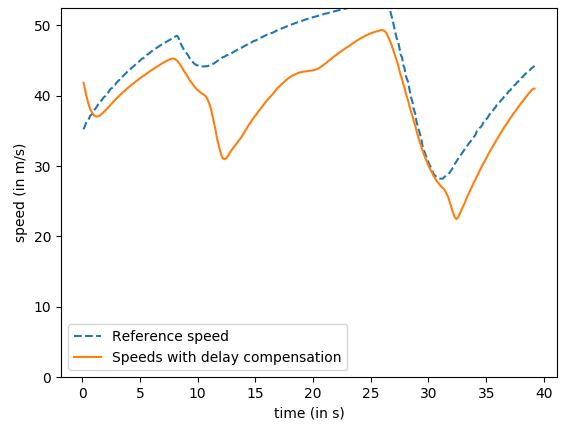}
    \caption{Ego vehicle speed for the case when delay compensation is considered}
    \label{fig:fig_speeds_with_exp1B}
\end{figure}

\begin{table}[htbp]
\begin{center}
\begin{tabular}{ m{2.5cm}  m{2cm}  m{2.5cm} } 
\toprule
 Run & Lap time (in s) & No of lane boundary violations \\ [0.5ex] 
 \midrule
 Without delay compensation & 38.04 & 2 \\
 With delay compensation & \textbf{36.51} & \textbf{0} \\
 \bottomrule
\end{tabular}
\end{center}
\caption{Comparison in lap times with and without delay compensation included}
\label{table:race_times_B}
\end{table}

\subsubsection{Overtaking at turn in racing scenario} \label{sec:exp2B}

\begin{figure}[htbp]
\begin{subfigure}{.5\textwidth}
    \centering
    \includegraphics[width=\textwidth]{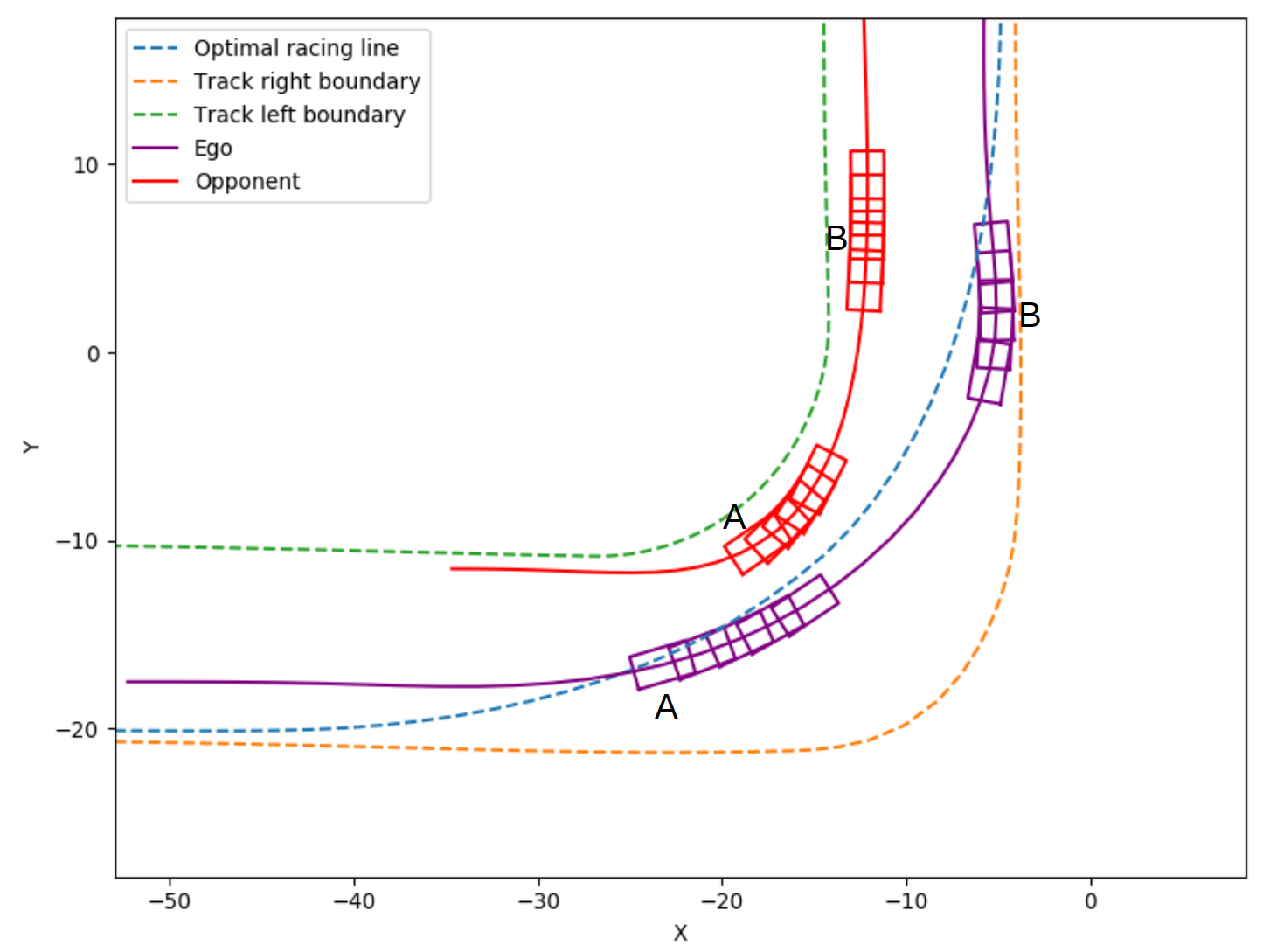}
    \caption{With delay compensation.}
    \label{fig:2B_with_comp}
\end{subfigure}
\newline
\begin{subfigure}{.5\textwidth}
    \centering
    \includegraphics[width=\textwidth]{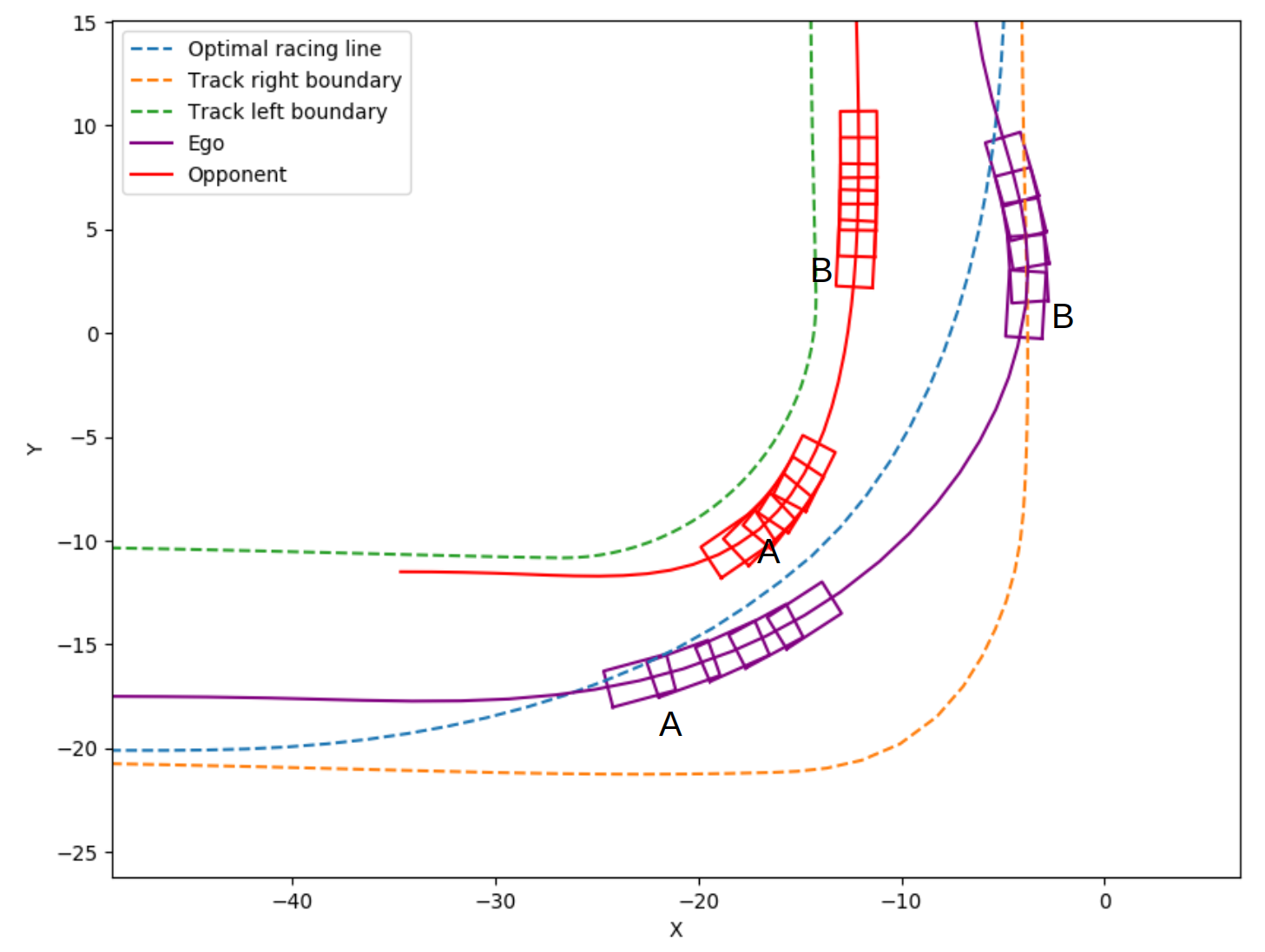}
    \caption{Without delay compensation.}
    \label{fig:2B_without_comp}
\end{subfigure}
\caption{Comparison in Experiment \ref{sec:exp2B}. The ego vehicle crashes at point A without any delay compensation.}
\end{figure}

Finally, to validate the controller plan A, we design a scenario where the vehicle has to overtake at a turn, forcing it to leave the optimal racing line at the turn. As the vehicle deviates from the optimal racing line to overtake and avoid collision, the vehicle has to apply braking to avoid moving out of the track. However, without the delay compensation (see Fig. \ref{fig:2B_without_comp}), the vehicle applies braking late, leading to the vehicle CBF constraint being unable to retain the vehicle within the lane boundaries. With delay compensation considered, the vehicle makes more effective control decisions taking the delay into account and thus is able to perform the overtaking maneuver safely at the turn (see Fig. \ref{fig:2B_with_comp}).

\section{Conclusion}
\label{sec:conclusions}

We propose a delay-aware robust tube MPC for safe autonomous driving and racing. We also compensate for delays of a learning-enabled controller trained in a delay-free environment and employed in the closed-loop system as a primary controller. The simulation results in the high-fidelity Gazebo and CARLA simulators show the efficiency and real-time performance of our designs.

% \section*{Acknowledgment}

% This material is based upon work supported by the United States Air Force and DARPA under Contract No. FA8750-18-C-0092. Any opinions, findings and conclusions or recommendations expressed in this material are those of the author(s) and do not necessarily reflect the views of the United States Air Force and DARPA.

\bibliographystyle{./bibliography/IEEEtran}
\bibliography{./bibliography/IEEEabrv,./bibliography/IEEEexample}

\begin{thebibliography}{10}
\providecommand{\url}[1]{#1}
\csname url@samestyle\endcsname
\providecommand{\newblock}{\relax}
\providecommand{\bibinfo}[2]{#2}
\providecommand{\BIBentrySTDinterwordspacing}{\spaceskip=0pt\relax}
\providecommand{\BIBentryALTinterwordstretchfactor}{4}
\providecommand{\BIBentryALTinterwordspacing}{\spaceskip=\fontdimen2\font plus
\BIBentryALTinterwordstretchfactor\fontdimen3\font minus
  \fontdimen4\font\relax}
\providecommand{\BIBforeignlanguage}[2]{{%
\expandafter\ifx\csname l@#1\endcsname\relax
\typeout{** WARNING: IEEEtran.bst: No hyphenation pattern has been}%
\typeout{** loaded for the language `#1'. Using the pattern for}%
\typeout{** the default language instead.}%
\else
\language=\csname l@#1\endcsname
\fi
#2}}
\providecommand{\BIBdecl}{\relax}
\BIBdecl

\bibitem{dvij2022}
\BIBentryALTinterwordspacing
D.~Kalaria, Q.~Lin, and J.~M. Dolan, ``Delay-aware robust control for safe
  autonomous driving,'' in \emph{33rd IEEE Intelligent Vehicles
  Symposium}.\hskip 1em plus 0.5em minus 0.4em\relax IEEE, 2022. [Online].
  Available: \url{https://arxiv.org/pdf/2109.07101.pdf}
\BIBentrySTDinterwordspacing

\bibitem{cortes2011delay}
P.~Cortes, J.~Rodriguez, C.~Silva, and A.~Flores, ``Delay compensation in model
  predictive current control of a three-phase inverter,'' \emph{IEEE
  Transactions on Industrial Electronics}, vol.~59, pp. 1323--1325, 2011.

\bibitem{su2013computation}
Y.~Su, K.~K. Tan, and T.~H. Lee, ``Computation delay compensation for real time
  implementation of robust model predictive control,'' \emph{Journal of Process
  Control}, vol.~23, no.~9, pp. 1342--1349, 2013.

\bibitem{nahidi2019study}
A.~Nahidi, A.~Khajepour, A.~Kasaeizadeh, S.-K. Chen, and B.~Litkouhi, ``A study
  on actuator delay compensation using predictive control technique with
  experimental verification,'' \emph{Mechatronics}, vol.~57, pp. 140--149,
  2019.

\bibitem{liao2018design}
Y.~Liao and F.~Liao, ``Design of preview controller for linear continuous-time
  systems with input delay,'' \emph{International Journal of Control,
  Automation and Systems}, vol.~16, no.~3, pp. 1080--1090, 2018.

\bibitem{6083022}
N.~E. Kahveci and P.~A. Ioannou, ``Automatic steering of vehicles subject to
  actuator saturation and delay,'' in \emph{2011 14th International IEEE
  Conference on Intelligent Transportation Systems (ITSC)}, 2011, pp. 119--124.

\bibitem{Peter2007}
P.~Wieland, ``Constructive safety using control barrier functions,'' vol.~7, 08
  2007, pp. 462--467.

\bibitem{Amescdc2014}
A.~D. {Ames}, J.~W. {Grizzle}, and P.~{Tabuada}, ``Control barrier function
  based quadratic programs with application to adaptive cruise control,'' in
  \emph{53rd IEEE Conference on Decision and Control}, 2014, pp. 6271--6278.

\bibitem{orosz2019safety}
G.~Orosz and A.~D. Ames, ``Safety functionals for time delay systems,'' in
  \emph{2019 American Control Conference (ACC)}.\hskip 1em plus 0.5em minus
  0.4em\relax IEEE, 2019, pp. 4374--4379.

\bibitem{singletary2020control}
A.~Singletary, Y.~Chen, and A.~D. Ames, ``Control barrier functions for
  sampled-data systems with input delays,'' in \emph{2020 59th IEEE Conference
  on Decision and Control (CDC)}.\hskip 1em plus 0.5em minus 0.4em\relax IEEE,
  2020, pp. 804--809.

\bibitem{abel2021safety}
I.~Abel, M.~Krsti{\'c}, and M.~Jankovi{\'c}, ``Safety-critical control of
  systems with time-varying input delay,'' \emph{IFAC-PapersOnLine}, vol.~54,
  no.~18, pp. 169--174, 2021.

\bibitem{feraco2020local}
S.~Feraco, S.~Luciani, A.~Bonfitto, N.~Amati, and A.~Tonoli, ``A local
  trajectory planning and control method for autonomous vehicles based on the
  rrt algorithm,'' in \emph{2020 AEIT International Conference of Electrical
  and Electronic Technologies for Automotive (AEIT AUTOMOTIVE)}.\hskip 1em plus
  0.5em minus 0.4em\relax IEEE, 2020, pp. 1--6.

\bibitem{stahl2019}
T.~Stahl, A.~Wischnewski, J.~Betz, and M.~Lienkamp, ``Multilayer graph-based
  trajectory planning for race vehicles in dynamic scenarios,'' in \emph{2019
  IEEE Intelligent Transportation Systems Conference (ITSC)}, 2019, pp.
  3149--3154.

\bibitem{tramacere2021local}
E.~Tramacere, S.~Luciani, S.~Feraco, S.~Circosta, I.~Khan, A.~Bonfitto, and
  N.~Amati, ``Local trajectory planning for autonomous racing vehicles based on
  the rapidly-exploring random tree algorithm,'' in \emph{International Design
  Engineering Technical Conferences and Computers and Information in
  Engineering Conference}, vol. 85369.\hskip 1em plus 0.5em minus 0.4em\relax
  American Society of Mechanical Engineers, 2021, p. V001T01A009.

\bibitem{Srinivasan2021AHM}
S.~Srinivasan, S.~N. Giles, and A.~Liniger, ``A holistic motion planning and
  control solution to challenge a professional racecar driver,'' \emph{IEEE
  Robotics and Automation Letters}, vol.~6, pp. 7854--7860, 2021.

\bibitem{Vzquez2020OptimizationBasedHM}
J.~L. V{\'a}zquez, M.~Br{\"u}hlmeier, A.~Liniger, A.~Rupenyan, and J.~Lygeros,
  ``Optimization-based hierarchical motion planning for autonomous racing,''
  \emph{2020 IEEE/RSJ International Conference on Intelligent Robots and
  Systems (IROS)}, pp. 2397--2403, 2020.

\bibitem{Kalaria2021LocalNO}
D.~Kalaria, P.~Maheshwari, A.~Jha, A.~K. Issar, D.~Chakravarty, S.~Anwar, and
  A.~Towar, ``Local nmpc on global optimised path for autonomous racing,''
  \emph{ArXiv}, vol. abs/2109.07105, 2021.

\bibitem{doi:10.1080/00423114.2019.1631455}
\BIBentryALTinterwordspacing
A.~Heilmeier, A.~Wischnewski, L.~Hermansdorfer, J.~Betz, M.~Lienkamp, and
  B.~Lohmann, ``Minimum curvature trajectory planning and control for an
  autonomous race car,'' \emph{Vehicle System Dynamics}, vol.~58, no.~10, pp.
  1497--1527, 2020. [Online]. Available:
  \url{https://doi.org/10.1080/00423114.2019.1631455}
\BIBentrySTDinterwordspacing

\bibitem{Beltman2008OptimizationOI}
F.~Beltman, ``Optimization of ideal racing line using discrete markov decision
  processes,'' 2008.

\bibitem{Rosolia2020LearningHT}
U.~Rosolia and F.~Borrelli, ``Learning how to autonomously race a car: A
  predictive control approach,'' \emph{IEEE Transactions on Control Systems
  Technology}, vol.~28, pp. 2713--2719, 2020.

\bibitem{Vesel2015RacingLO}
R.~W. Vesel, ``Racing line optimization @ race optimal,'' \emph{ACM
  SIGEVOlution}, vol.~7, pp. 12 -- 20, 2015.

\bibitem{Jain2020ComputingTR}
A.~Jain and M.~Morari, ``Computing the racing line using bayesian
  optimization,'' \emph{2020 59th IEEE Conference on Decision and Control
  (CDC)}, pp. 6192--6197, 2020.

\bibitem{Werling2010OptimalTG}
M.~Werling, J.~Ziegler, S.~Kammel, and S.~Thrun, ``Optimal trajectory
  generation for dynamic street scenarios in a fren{\'e}t frame,'' \emph{2010
  IEEE International Conference on Robotics and Automation}, pp. 987--993,
  2010.

\bibitem{Deo2018ConvolutionalSP}
N.~Deo and M.~M. Trivedi, ``Convolutional social pooling for vehicle trajectory
  prediction,'' \emph{2018 IEEE/CVF Conference on Computer Vision and Pattern
  Recognition Workshops (CVPRW)}, pp. 1549--15\,498, 2018.

\bibitem{pan2020safe}
Y.~Pan, Q.~Lin, H.~Shah, and J.~M. Dolan, ``Safe planning for self-driving via
  adaptive constrained ilqr,'' in \emph{2020 IEEE/RSJ International Conference
  on Intelligent Robots and Systems (IROS)}.\hskip 1em plus 0.5em minus
  0.4em\relax IEEE, 2020, pp. 2377--2383.

\bibitem{khaitan2021safe}
S.~Khaitan, Q.~Lin, and J.~M. Dolan, ``Safe planning and control under
  uncertainty for self-driving,'' \emph{IEEE Transactions on Vehicular
  Technology}, vol.~70, no.~10, pp. 9826--9837, 2021.

\bibitem{book}
R.~Rajamani, \emph{Vehicle Dynamics and Control}, 01 2006.

\bibitem{Zeng2021SafetyCriticalMP}
J.~Zeng, B.~Zhang, and K.~Sreenath, ``Safety-critical model predictive control
  with discrete-time control barrier function,'' \emph{2021 American Control
  Conference (ACC)}, pp. 3882--3889, 2021.

\bibitem{WILLS20041415}
\BIBentryALTinterwordspacing
A.~G. Wills and W.~P. Heath, ``Barrier function based model predictive
  control,'' \emph{Automatica}, vol.~40, no.~8, pp. 1415--1422, 2004. [Online].
  Available:
  \url{https://www.sciencedirect.com/science/article/pii/S0005109804000809}
\BIBentrySTDinterwordspacing

\bibitem{1383612}
R.~Smith, ``Robust model predictive control of constrained linear systems,'' in
  \emph{Proceedings of the 2004 American Control Conference}, vol.~1, 2004, pp.
  245--250 vol.1.

\bibitem{gao2014}
Y.~Gao, A.~Gray, H.~E. Tseng, and F.~Borrelli, ``A tube-based robust nonlinear
  predictive control approach to semiautonomous ground vehicles,''
  \emph{Vehicle System Dynamics}, vol.~52, no.~6, pp. 802--823, 2014.

\bibitem{article}
S.~V. Raković, E.~Kerrigan, K.~Kouramas, and D.~Mayne, \emph{Invariant
  approximations of robustly positively invariant sets for constrained linear
  discrete-time systems subject to bounded disturbances}.\hskip 1em plus 0.5em
  minus 0.4em\relax University of Cambridge, Department of Engineering
  Cambridge, 2004.

\bibitem{article_kalman}
I.~Hashlamon and K.~Erbatur, ``An improved real-time adaptive kalman filter
  with recursive noise covariance updating rules,'' \emph{Turkish Journal of
  Electrical Engineering and Computer Sciences}, 12 2013.

\bibitem{myers1976adaptive}
K.~Myers and B.~Tapley, ``Adaptive sequential estimation with unknown noise
  statistics,'' \emph{IEEE Transactions on Automatic Control}, vol.~21, no.~4,
  pp. 520--523, 1976.

\bibitem{liu2015safe}
C.~Liu and M.~Tomizuka, ``Safe exploration: Addressing various uncertainty
  levels in human robot interactions,'' in \emph{2015 American Control
  Conference (ACC)}.\hskip 1em plus 0.5em minus 0.4em\relax IEEE, 2015, pp.
  465--470.

\bibitem{koenig2004design}
N.~Koenig and A.~Howard, ``Design and use paradigms for gazebo, an open-source
  multi-robot simulator,'' in \emph{2004 IEEE/RSJ International Conference on
  Intelligent Robots and Systems (IROS)(IEEE Cat. No. 04CH37566)},
  vol.~3.\hskip 1em plus 0.5em minus 0.4em\relax IEEE, 2004, pp. 2149--2154.

\bibitem{dosovitskiy2017carla}
A.~Dosovitskiy, G.~Ros, F.~Codevilla, A.~Lopez, and V.~Koltun, ``Carla: An open
  urban driving simulator,'' in \emph{Conference on robot learning}.\hskip 1em
  plus 0.5em minus 0.4em\relax PMLR, 2017, pp. 1--16.

\bibitem{claviere2019trajectory}
A.~Claviere, S.~Dutta, and S.~Sankaranarayanan, ``Trajectory tracking control
  for robotic vehicles using counterexample guided training of neural
  networks,'' in \emph{Proceedings of the International Conference on Automated
  Planning and Scheduling}, vol.~29, 2019, pp. 680--688.

\end{thebibliography}

\appendix

\section*{Full expansion for obstacle control barrier function}
\label{App:collision_constraints}

\begin{figure}[htbp]
    \centering
    \includegraphics[width=0.4\textwidth]{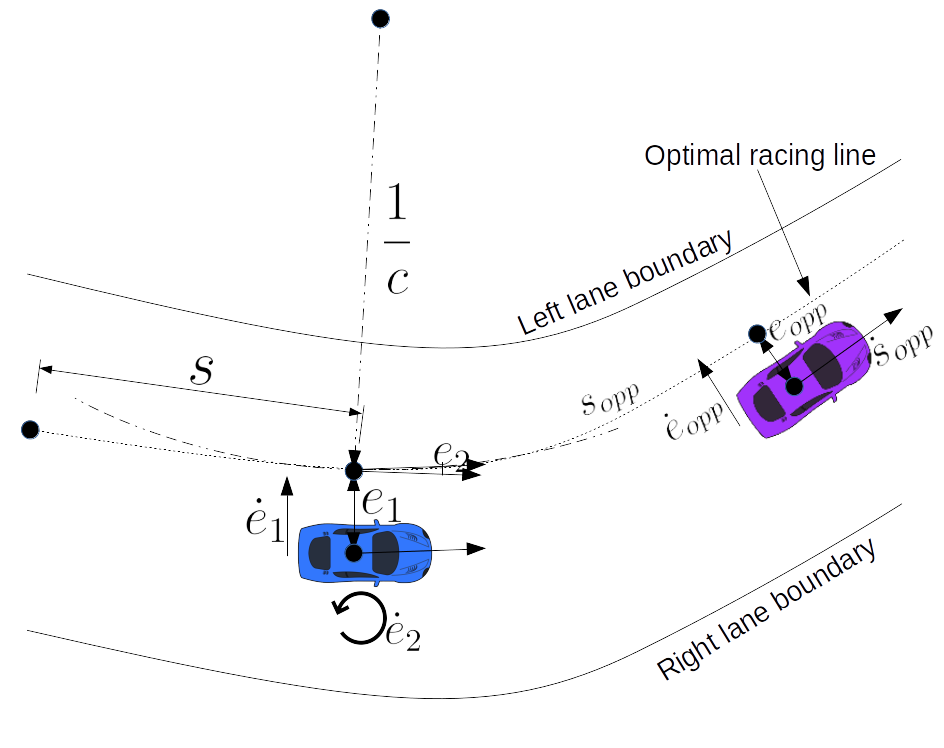}
    \caption{Illustration for the derivation of obstacle control barrier function}
    \label{fig:fig_control_barrier}
\end{figure}

We define the safety function $h(\cdot)$ as follows:
\begin{equation}
    h(\mathbf{x}, \mathbf{x}_{opp}) = \left(\frac{s-s_{opp}}{d_s}\right)^2 + \left(\frac{e_1 - e_{opp}}{d_e}\right)^2 - 1
\end{equation} 
where $s, s_{opp}, e_1, e_{opp}$ defined before are illustrated in Fig. \ref{fig:fig_control_barrier}. $d_s$ and $d_e$ are the longitudinal and lateral safe distances from the target opponent vehicle, $\mathbf{x}$ and $\mathbf{x}_{opp}$ are the states of the ego vehicle and the opponent vehicle. We reformulate the vehicle dynamic model (see Eq. \ref{eq:dynamic_eqn}) in a control affline form as follows:
\begin{equation}
    \begin{split}
    &\dot{\mathbf{x}} = f(\mathbf{x}) + g(\mathbf{x}) \mathbf{u}\\
    &\text{with } f(\mathbf{x}) = \\
            &\begin{bmatrix}
            \dot e_1 \\
            -\frac{2 C_f + 2 C_r}{m v_x} \dot e_1 + \frac{2 C_f + 2 C_r}{m} e_2 - \frac{-2 C_f l_f + 2 C_r l_r}{m v_x} \dot e_2 \\
            \dot e_2 \\
            -\frac{2 C_f l_f - 2 C_r l_r}{I_z v_x} \dot e_1 + \frac{2 C_f l_f - 2 C_r l_r}{I_z} e_2 - \frac{2 C_f l_f^2 + 2 C_r l_r^2}{I_z v_x} \dot e_2 \\
            - K_f v_x\\
            v_x - \dot e_1 e_2\\
            -K_{\delta} \delta_a
        \end{bmatrix}\\
        &+
        \begin{bmatrix}
            0 \\
            - v_x (v_x + \frac{2 C_f l_f - 2 C_r l_r}{m v_x}) c + \frac{4 C_f}{m} \delta_a\\
            0 \\
            - \frac{2 C_f l_f^2 + 2 C_r l_r^2}{I_z} c\\
            0\\
            0\\
            0
        \end{bmatrix} \\
        & g(\mathbf{x}) = \begin{bmatrix}
            0 \ \ \ 0 \\
            0 \ \ \ 0\\
            0  \ \ \ 0\\
            0  \ \ \ 0\\
            0  \ \ K_p\\
            0  \ \ \ 0\\
            K_{\delta}  \ \ \ 0 
        \end{bmatrix} 
        \begin{bmatrix}
            \delta \\
            p
        \end{bmatrix}
    \end{split}
\end{equation}

The first order CBF requires $\dot{h}(\mathbf{x}, \mathbf{x}_{opp}) + \lambda h(\mathbf{x}, \mathbf{x}_{opp}) \ge 0$. By the chain rule, we have $\Delta_\mathbf{x} h(\cdot)\dot{\mathbf{x}}+ \Delta_{\mathbf{x}_{opp}} h(\cdot)\dot{\mathbf{x}_{opp}} + \lambda h(\mathbf{\cdot}) \ge 0$, i.e.,
\begin{equation}
    \begin{split}
    &\Delta_x h(\cdot) [f(\mathbf{x}) + g(\mathbf{x}) \mathbf{u}] + \Delta_{\mathbf{x}_{opp}} h(\cdot)\dot{\mathbf{x}_{opp}} + \lambda h(\cdot) \ge 0\\
    &\text{thus,} \left[\frac{2(e_1-e_{opp})}{d_e} \  0 \  0 \  0 \  0 \  \frac{2(s-s_{opp})}{d_s} \  0\right] (f(\mathbf{x}) + g(\mathbf{x}) \mathbf{u}) \\
    & -\frac{2(e_1-e_{opp})}{d_e} \dot{e_{opp}}-\frac{2(s-s_{opp})}{d_s} \dot{s_{opp}}\\
    &+ \lambda \left(\left(\frac{s-s_{opp}}{d_s}\right)^2 + \left(\frac{e_1 - e_{opp}}{d_e}\right)^2 - 1\right) \ge 0 \\
\end{split}
\end{equation}
where the Lie derivitive $L_g h(x)\coloneqq \Delta_x h(\mathbf(x)) g(\mathbf{x})=0$. It implies that using the first order CBF doesn't directly contain any constraint on the control commands $\mathbf{u}$. We continue taking derivative until the constraints of the control appear. We use the second-order CBF by defining $\phi(\mathbf{x}, \mathbf{x}_{opp}) = \dot{h}(\mathbf{x}, \mathbf{x}_{opp}) + \lambda h(\mathbf{x}, \mathbf{x}_{opp})$. For the new safety constraint $\dot{\phi}(\cdot) + \phi(\cdot) \ge 0$, we have

\begin{equation}
\begin{split}
&\ddot{h}(\cdot) + \lambda \dot{h}(\cdot) + \lambda(\dot{h}(\cdot) + \lambda h(\cdot)) \ge 0\\
&\text{i.e., } \ddot{h}(\cdot) + 2 \lambda \dot{h}(\cdot) + \lambda^2 h(\cdot) \ge 0\\
&\text{By expanding, we get}\\
% &2 \frac{(v_x - \dot e_1 e_2 - \dot{s}_{opp})^2}{d_s} + 2 \frac{(\dot{e_1} - \dot{e_{opp}})^2}{d_e} \\
% &+ 2 \frac{(s - s_{opp}) (\dot{v_x}-\ddot{e}_1 e_2 -\dot{e}_1 \dot{e}_2)}{d_s} \\
% &- 2 \frac{(e_1-e_{opp})}{d_s} (\frac{2 C_f + 2 C_r}{m V_x} \dot e_1 + \frac{2 C_f + 2 C_r}{m} e_2 \\
% &- \frac{-2 C_f l_f + 2 C_r l_r}{m V_x} \dot e_2) + 2 \lambda \dot{h}(\mathbf{x}) + h(\mathbf{x}) \ge 0\\
&2 \frac{(v_x - \dot e_1 e_2 - \dot{s}_{opp})^2}{d_s} + 2 \frac{(\dot{e_1} - \dot{e_{opp}})^2}{d_e} \\
&+ 2 \frac{(s - s_{opp})}{d_s} (K_p p - K_f v_x -\left(-\frac{2 C_f + 2 C_r}{m v_x} \dot e_1 \right. \\
& \left. + \frac{2 C_f + 2 C_r}{m} e_2 - v_x \left(v_x + \frac{2 C_f l_f - 2 C_r l_r}{m v_x}\right) c + \frac{4 C_f}{m} \delta \right) e_2 \\
&-\dot{e}_1 \dot{e}_2) - 2 \frac{(e_1-e_{opp})}{d_s} \left(\frac{2 C_f + 2 C_r}{m V_x} \dot e_1 + \frac{2 C_f + 2 C_r}{m} e_2 \right.\\
& \left.- \frac{-2 C_f l_f + 2 C_r l_r}{m V_x} \dot e_2  v_x \left(v_x + \frac{2 C_f l_f - 2 C_r l_r}{m v_x}\right) c + \frac{4 C_f}{m} \delta \right) \\
&+ 2 \lambda \left(2 \frac{(s - s_{opp})}{d_s} (v_x - \dot e_1 e_2 - \dot{s}_{opp}\right) \\
&+ 2 \frac{(e_1 - e_{opp})}{d_e} (\dot{e_1} - \dot{e_{opp}})) \\
&+\lambda^2 \left(\left(\frac{s-s_{opp}}{d_s}\right)^2 + \left(\frac{e_1 - e_{opp}}{d_e}\right)^2 - 1\right) \ge 0\\
\end{split}
\end{equation}

It can be observed that the terms in front of the control actions are non-negative. Therefore, we can use a second-order CBF for this system with the relative degree of two.  
\end{document}